\documentclass{article} % For LaTeX2e
\usepackage{iclr2023_conference,times}

\usepackage{amsmath,amsfonts,bm}

\def\figref#1{figure~\ref{#1}}
\def\eqref#1{equation~\ref{#1}}
\def\1{\bm{1}}

\DeclareMathAlphabet{\mathsfit}{\encodingdefault}{\sfdefault}{m}{sl}
\SetMathAlphabet{\mathsfit}{bold}{\encodingdefault}{\sfdefault}{bx}{n}

\usepackage{bm}
\usepackage{etoc}

\usepackage{hyperref}
\usepackage{url}
\usepackage{microtype}
\usepackage{graphicx}
\usepackage{booktabs} 
\usepackage{caption}
\usepackage{wrapfig}
\usepackage{subcaption}
\usepackage{microtype}
\usepackage{graphicx}

\usepackage{xcolor}
\definecolor{mintbg}{rgb}{.63,.79,.95}
\usepackage{todonotes} 
\usepackage{amssymb}
\colorlet{lightmintbg}{mintbg!40}

\newcommand{\calD}{\mathcal D}
\newcommand{\calG}{\mathcal G}
\newcommand{\calH}{$\mathcal H$}
\newcommand{\calM}{\mathcal M}
\newcommand{\pa}{{\text{pa}}}

\newcommand{\cpd}{CPD}
\newcommand{\cpds}{CPDs}
\renewcommand{\eqref}[1]{Eq. (\ref{#1})}
\renewcommand{\figref}[1]{Fig. \ref{#1}}
\newcommand{\T}{\mathsf{T}}
\newcommand{\CSIVA}{CSIvA}

\title{Learning to Induce Causal Structure}

\author{
Nan Rosemary Ke \textsuperscript{1,2,3},
Silvia Chiappa \textsuperscript{1},
Jane Wang \textsuperscript{1},
{Anirudh  Goyal} \textsuperscript{1,2,4} \\
\textbf{Jorg Bornschein} \textsuperscript{1},
\textbf{Melanie Rey} \textsuperscript{1},
\textbf{Theophane Weber} \textsuperscript{1},
\textbf{Matthew Botvinick} \textsuperscript{1}\\
\textbf{Michael Mozer} \textsuperscript{5},
\textbf{Danilo Jimenez Rezende} \textsuperscript{1}}

\iclrfinalcopy % Uncomment for camera-ready version, but NOT for submission.
\begin{document}

\let\footnote\relax\footnotetext{\textsuperscript{1}DeepMind, \textsuperscript{2} Mila, \textsuperscript{3} Polytechnique Montreal,  \textsuperscript{4} University of Montreal, \textsuperscript{5} Google Research, Brain Team, %\textsuperscript{*} contributed during internship at DeepMind.
Corresponding author: \texttt{nke@google.com}
}
\maketitle

\begin{abstract}
The fundamental challenge in causal induction is to infer the underlying graph structure given observational and/or interventional data. Most existing causal induction algorithms operate by generating candidate graphs and evaluating them using either score-based methods (including continuous optimization) or independence tests. In our work, 
we instead treat the inference process as a black box and design a neural network architecture that learns the mapping from \emph{both observational and interventional data} to graph structures via supervised training on synthetic graphs. The learned model generalizes to new synthetic graphs, is robust to train-test distribution shifts, and achieves state-of-the-art performance on naturalistic graphs for low sample complexity.
\end{abstract}

\section{Introduction}
\label{intro}
\begin{figure}
\centering
%\begin{subfigure}[b]{1\linewidth}
%{
\vspace{-1\baselineskip}
\includegraphics[width=0.90\linewidth]{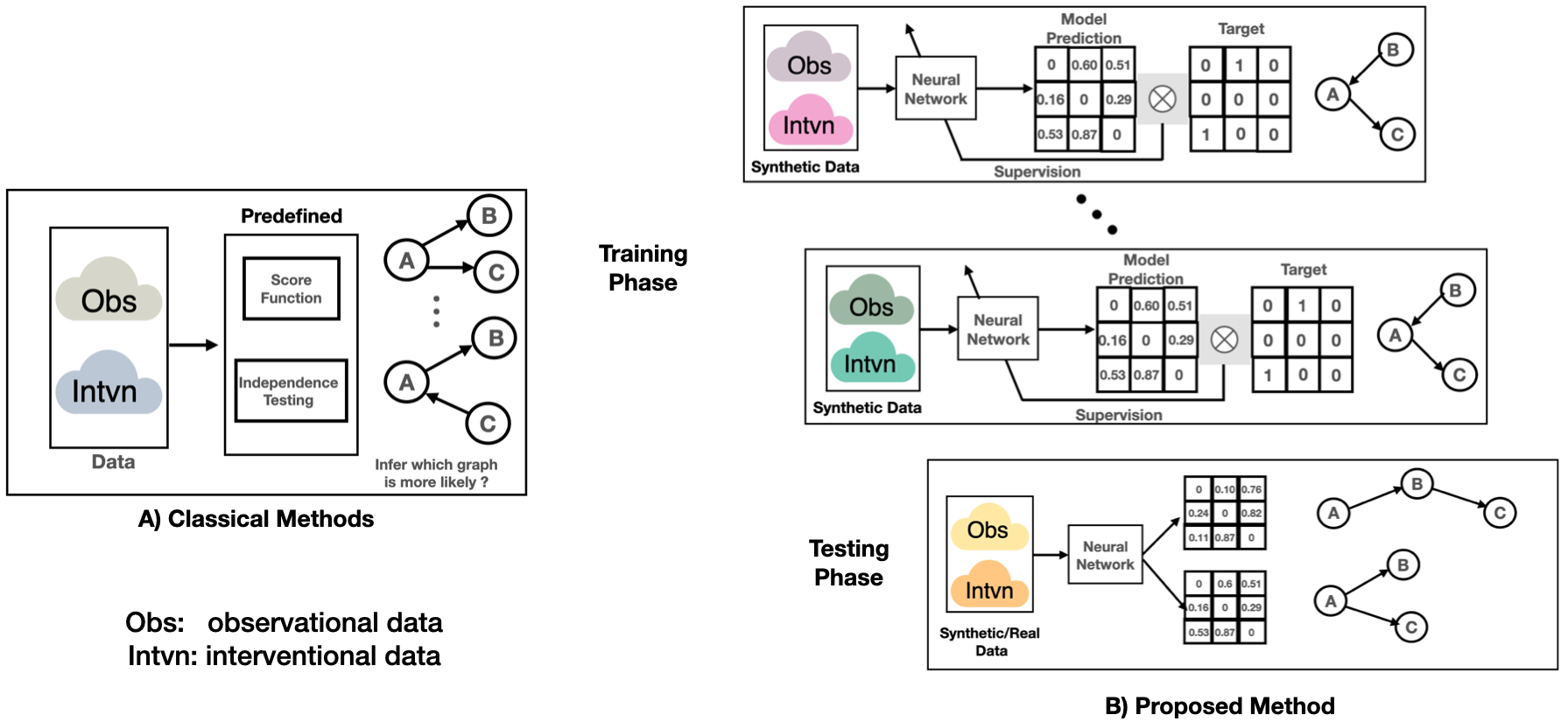}
\caption{(A). Standard unsupervised approach to causal structure induction: Algorithms use a predefined scoring metric or statistical independence tests to select the best candidate structures. (B). Our supervised approach to causal structure induction (via attention; \CSIVA): A model is presented with data and structures as training pairs and learns a mapping between them.}
\label{fig:(Un)Supervised}
\end{figure}

The problem of discovering the causal relationships that govern a system through observing its behavior, either passively (\textit{observational data}) or by manipulating some of its variables (\textit{interventional data}), lies at the core of  many scientific disciplines, including medicine, biology, and economics. By using the graphical formalism of causal Bayesian networks (CBNs) \citep{kollerl2009probabilistic,pearl2009causality}, this problem can be framed as inducing the graph structure that best represents the relationships. Most approaches to causal structure induction are based on an unsupervised learning paradigm in which the structure is directly inferred from the system observations, either by ranking different structures according to some metrics (score-based approaches) or by determining the presence of an edge between pairs of variables using conditional independence tests (constraint-based approaches) \citep{drton2017structure,glymor2019review,heinze2018causal,heinze2018invariant} (see Fig.\ \ref{fig:(Un)Supervised}(a)). The unsupervised paradigm poses however some challenges: score-based approaches are burdened with the high computational cost of having to explicitly consider all possible structures 
%which grows super-exponentially with the number of variables, 
and with the difficulty of devising metrics that can balance goodness of fit with constraints for differentiating causal from purely statistical relationships (e.g. sparsity of the structure or simplicity of the generation mechanism); constraint-based methods are sensitive to failure of independence tests and require faithfulness, a property that does not hold in many real-world scenarios \citep{koski2012review,mabrouk2014efficient}.

Recently, supervised learning methods based on observational data have been introduced as an alternative to unsupervised  approaches \citep{lopez2015randomized,lopez2015towards,li2020supervised}. In this work, we extend the supervised learning paradigm to also use interventional data, enabling greater flexibility. We propose
%where %we propose a supervised learning paradigm in which 
a model that is first trained on synthetic data generated using different CBNs to learn a mapping from data to graph structures and then used to induce the structures underlying datasets of interest (see Fig.\ \ref{fig:(Un)Supervised}(b)).
The model is a novel variant of a transformer neural network that receives as input a dataset consisting of observational and interventional samples corresponding to the same CBN and outputs a prediction of the CBN graph structure. The mapping from the dataset to the underlying structure is achieved through an attention mechanism which alternates between attending to different variables in the graph and to different samples from a variable. The output is produced by a decoder mechanism that operates as an autoregressive generative model on the inferred structure. Our approach can be viewed as a form of meta-learning, as the model learns about the relationship between datasets and structures underlying them. 
%By allowing the use of \emph{both} observational \emph{and} \textit{interventional} data, our method enables greater flexibility. 

A requirement of a supervised approach would seem to be that the distributions of the training and test data match or highly overlap. Obtaining real-world training data with a known causal structure that matches test data from multiple domains is extremely challenging. We show that meta-learning enables the model to generalize well to data from naturalistic CBNs even if trained on synthetic data with relatively few assumptions. We show that our model can learn a mapping from datasets to structures and achieves state-of-the-art performance on classic benchmarks such as the Sachs, Asia and Child datasets \citep{lauritzen1988local,sachs2005causal,Spiegelhalter1992Learning}, despite never directly being trained on such data. Our contributions can be summarized as follows:
{\setlength{\leftmargini}{10pt}
\begin{itemize}%[leftmargin=5.5mm]
\setlength{\itemsep}{7pt}  
\setlength{\parskip}{-4pt}
\setlength{\parsep}{-4pt}
\item We tackle causal structure induction with a supervised approach (\CSIVA) that maps datasets composed of \textit{both observational and interventional} samples to structures.
%\item We introduce a novel variant a transformer neural network with an attention mechanism designed to discover relationships among variables %\silvia{imcomplete add 'and samples from a variable?'}.
\item We introduce a variant of a transformer architecture whose attention mechanism is structured to discover relationships among variables across samples.
\item We show that \CSIVA~ generalizes to novel structures, whether or not training and test distributions match. Most importantly, training on synthetic data transfers effectively to naturalistic CBNs.
\item We show that \CSIVA~ significantly outperforms state-of-the-art  causal discovery methods such as DCDI~\citep{brouillard2020differentiable}, ENCO~\citep{lippe2021efficient} both on various types of synthetic CBNs, as well as  on naturalistic CBNs. %With thorough ablations, we show the usefuless of various different components of \CSIVA.
\end{itemize}}
\section{Background}
\label{sec:background}
In this section we give some background on causal Bayesian networks and transformer neural networks, which form the main ingredients of our approach (see Appendix \ref{sec:app-transformers} for more details).

\textbf{Causal Bayesian networks (CBNs).}  A \emph{causal Bayesian network} \citep{kollerl2009probabilistic,pearl2009causality} is a pair $\calM=\langle\calG, p\rangle$, where $\calG$ is a \emph{directed acyclic graph} (DAG) whose nodes $X_1,\ldots, X_N$ represent random variables and edges express casual dependencies among them, and where $p$ is a joint distribution over all nodes such that $p(X_1, \ldots, X_N) = \prod_{n=1}^N p(X_n \,|\, \pa(X_n))$, where $\pa(X_n)$ are the \emph{parents} of $X_n$, i.e. the nodes with an edge onto $X_n$ (direct causes of $X_n$).
An input to the transformer neural network is formed by a dataset $\calD=\{x^s\}_{s=1}^S$, where $x^s:=(x^s_1,\ldots, x^s_N)^\T$ is either an  \emph{observational data sample} or an \emph{interventional data sample} obtained by performing an intervention on a randomly selected node in $\calG$. 
Observational data samples are samples from $p(X_1, \ldots, X_N)$. Except where otherwise noted, for all experimental settings, we considered \emph{hard interventions} on a node $X_n$ that consist in replacing the conditional probability distribution (\cpd) $p(X_n \,|\, \pa(X_n))$ with a delta function $\delta_{X_{n'}=x}$ forcing $X_{n'}$ to take on value $x$. Additional experiments were also performed using \emph{soft interventions}, which consisted of replacing $p(X_n \,|\, \pa(X_n))$ with a different \cpd~ $p'(X_n \,|\, \pa(X_n))$. An \emph{interventional data sample} is a sample from 
$\delta_{X_{n'}=x}\prod_{n=1,n\neq n'}^N p(X_n \,|\, \pa(X_n))$ 
in the first case, and a sample from $p'(X_n \,|\, \pa(X_n))\prod_{n=1,n\neq n'}^N p(X_n \,|\, \pa(X_n))$ in the second case. The structure of $\calG$ can be represented by an adjacency matrix $A$, defined by setting the $(k,l)$ entry, $A_{k,l}$, to 1 if there is an edge from $X_l$ to $X_k$ and to 0 otherwise. Therefore, 
the $n$-th row of $A$, denoted by $A_{n,:}$, indicates the parents of $X_n$
while the $n$-th column, denoted by $A_{:,n}$, indicates the \textit{children} of $X_n$.

\textbf{Transformer neural network.} 
A transformer \citep{devlin2018bert,vaswani2017attention} is a neural network equipped with layers of self-attention that make them suited to modelling structured data. In traditional applications, attention is used to account for the sequentially ordered nature of the data, e.g. modeling a sentence as a stream of words. In our case, each input of the transformer is a dataset of observational or interventional samples corresponding to the same CBN. Attention is thus used to account for the structure induced by the CBN graph structure and by having different samples from the same node. Transformers are permutation invariant with respect to the positions of the input elements, ensuring that the graph structure prediction does not depend on node and sample position.
\section{Causal Structure Induction via Attention (\CSIVA)}
\begin{wrapfigure}[19]{r}{0.65\textwidth}
\vspace{-1.7\baselineskip}
    \begin{center}
    \includegraphics[scale=0.13]{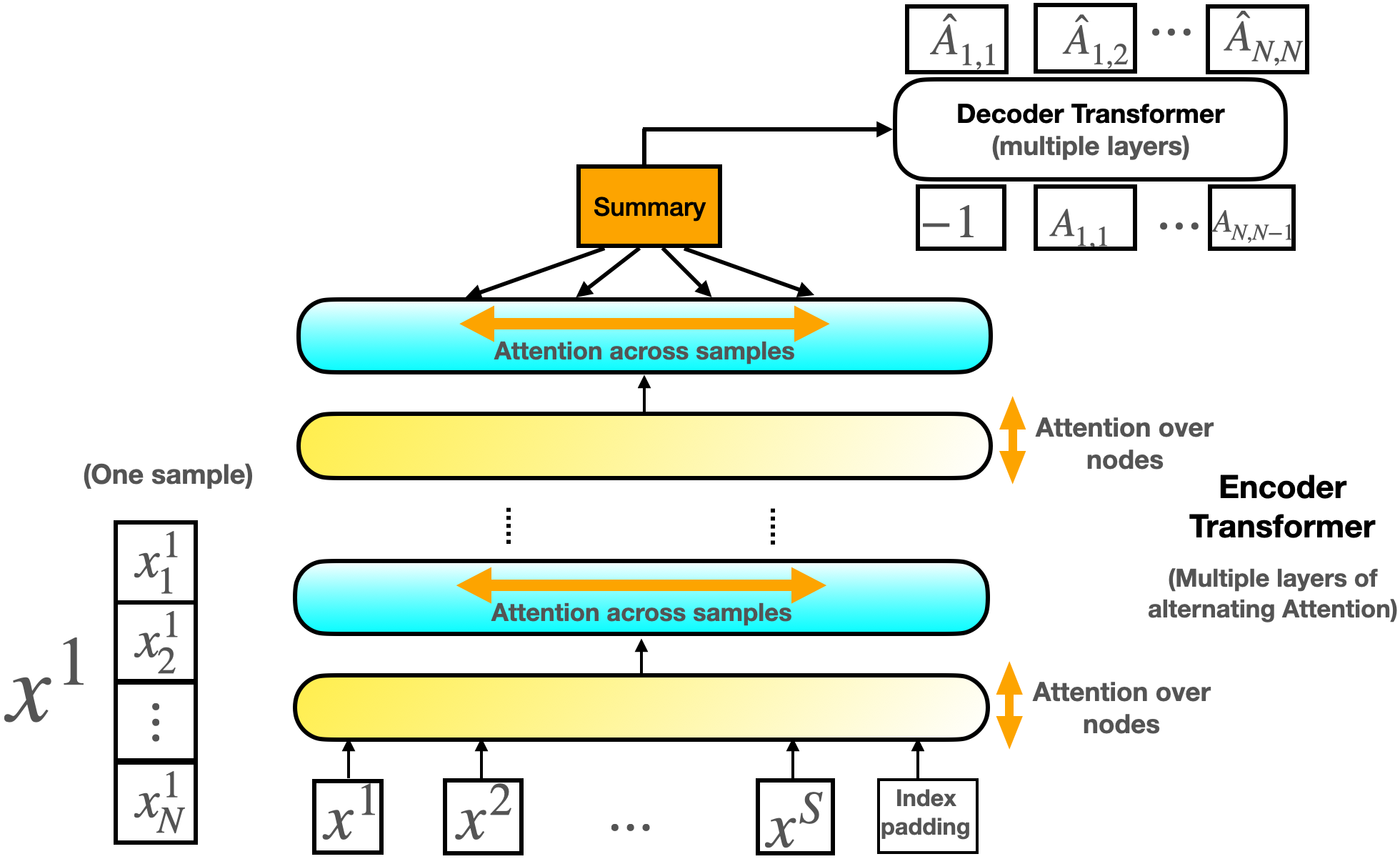}
    \end{center}
    \caption{Our model architecture and the structure of the input and output at training time. The input is a dataset $\calD=\{x^s:=(x^s_1,\ldots, x^s_N)^\T\}_{s=1}^S$ of $S$ samples from a CBN and its adjacency matrix $A$. The output is a prediction $\hat A$ of $A$. 
    }
    \vspace{.9\baselineskip}
    \label{fig:main_fig}
\end{wrapfigure}
Our approach is to treat causal structure induction as a supervised learning problem, by training a neural network to learn to map \emph{observational and interventional} data to the graph structure of the underlying CBN. Obtaining diverse, real-world data with known causal relationships in amounts sufficient for supervised training is not feasible. 
The key contribution of this work is to introduce a method that uses synthetic data generated from CBNs with different graph structures and associated \cpds~that is robust to shifts between the training and test data distributions. 
\vspace{-1mm}
\subsection{Supervised approach}
\vspace{-1mm}
We learn a distribution of graphs conditioned on observational and interventional data as follows. 
We generate training data from a joint distribution $t(\mathcal{G}, \mathcal{D})$, between a graph $\mathcal{G}$ and a dataset $\mathcal{D}$ comprising of $S$ \textit{observational} and \textit{interventional} samples from a CBN associated to $\mathcal{G}$ as follows. We first sample a set of graphs $\{\calG^i\}_{i=1}^I$ with nodes $X^i_1,\ldots,X^i_N$ from a common distribution $t(\mathcal{G})$ as described in Section \ref{sec:StructureDistribution} (to simplify notation, in the remainder of the paper we omit the graph index $i$ when referring to nodes), and then associate random \cpds~to the graphs as described in Section \ref{sec:CPDs}.
This results in a set of CBNs $\{\calM^i\}_{i=1}^I$. For each CBN $\calM^i$, we then create a dataset $\calD^i=\{x^s\}_{s=1}^S$, where each element $x^s:=(x^s_1,\ldots, x^s_N)^\T$ is either an observational data sample or an \textit{interventional} data sample obtained by performing an intervention on a randomly selected node in $\calG^i$.

Our model defines a distribution $\hat{t}(\mathcal{G} \,|\, \mathcal{D}; \Theta)$ over graphs conditioned on observational and \textit{interventional} data and parametrized by $\Theta$. Specifically, 
$\hat{t}(A \,|\, \mathcal{D}; \Theta)$ has the following auto-regressive form: $\hat{t}(A \,|\, \mathcal{D}; \Theta) = \prod_{l=1}^{N^2} \sigma(A_{l}; \hat{A}_l = f_{\Theta}(A_{1, \ldots, (l-1)}, \mathcal{D}))$, where $\sigma(\cdot; \rho)$ is the Bernoulli distribution with parameter $\rho$, which is a function $f_{\Theta}$  built from an encoder-decoder architecture explained in Section \ref{sec:model_arcitecture} taking as input previous elements of the adjacency matrix $A$ (represented here as an array of $N^2$ elements) and $\mathcal{D}$.
% \subsubsection{Model training}\label{sec.training}
It is trained via maximum likelihood estimation (MLE), i.e $\Theta^* = \text{argmin}_{\Theta} \mathcal{L}(\Theta)$, where $\mathcal{L}(\Theta)= -\mathbb{E}_{(\mathcal{G}, \mathcal{D}) \sim t}[\ln \hat{t}(\mathcal{G} \,|\, \mathcal{D}; \Theta)]$, which corresponds to the usual cross-entropy (CE) loss for the Bernoulli distribution. Training is achieved using a stochastic gradient descent (SGD) approach in which each gradient update is performed using a pair $(\calD^i, A^i)$. %The procedure is summarized in Algorithm \ref{alg:supervised}. 
The data-sampling distribution $t(\mathcal{G}, \mathcal{D})$ and the MLE objective uniquely determine the target distribution learned by the model. In the infinite capacity case, $\hat{t}(\cdot \,|\, \mathcal{D}; \Theta^*) = t(\cdot \,|\, \mathcal{D})$. To see this, it suffices to note that the MLE objective $\mathcal{L}(\Theta)$ can be written as $\mathcal{L}(\Theta) = 
\mathbb{E}_{\mathcal{D} \sim t}[
\text{KL}(\hat{t}(\cdot \,|\, \mathcal{D}; \Theta); t(\cdot \,|\, \mathcal{D}))
] + c$, where KL is the Kullback-Leibler divergence and $c$ is a constant. In the finite-capacity case, the distribution defined by the model $\hat{t}(\cdot \,|\, \mathcal{D}; \Theta^*)$ is only an approximation of $t(\cdot \,|\, \mathcal{D})$.
\subsection{Model architecture}
\label{sec:model_arcitecture}
The function $f_{\Theta}$ defining the model's probabilities is built using two transformer networks. It is formed by an encoder transformer and by a decoder transformer (which we refer to as ``encoder'' and ``decoder'' for short). 
At training time, 
the encoder receives as input dataset $\calD^i$ 
and outputs a representation that summarizes the relationship between nodes in $\calG^i$. The decoder then recursively outputs predictions of the elements of the adjacency matrix $A^i$ using as input the elements previously predicted and the encoder output. This is shown in \figref{fig:main_fig} (where we omitted index $i$, as in the remainder of the section). At test time we obtain deterministic predictions of the adjacency matrix elements by taking the argmax of the Bernoulli distribution for use as inputs to the decoder.
\subsubsection{Encoder}
Our encoder is structured as an $(N+1) \times (S+1)$ lattice. The $N \times S$ part of the lattice 
formed by the first $N$ rows and first $S$ columns receives a dataset $\calD=\{(x^s_1,\ldots, x^s_N)^\T\}_{s=1}^S$. 
This is unlike standard transformers which typically receive as input a single data sample (e.g., a sequence of words in neural machine
translation applications) rather than a set of data samples.
Row $N+1$ of the lattice is used to specify whether each
data sample is \emph{observational}, through value $-1$, or \textit{interventional}, through integer value in $\{1,\ldots,N\}$ to
indicate the intervened node.

The goal of the encoder is to infer causal relationships between nodes by examining the set of samples. The transformer performs this inference in multiple stages, each represented by one transformer layer, 
such that each layer yields a $(N+1) \times (S+1)$ lattice of representations. The transformer
is designed to deposit its summary representation of the causal structure in column $S+1$.

\textbf{Embedding of the input.}
Each data-sample element $x^s_n$ is embedded into a vector of dimensionality $H$. 
Half of this vector is allocated to embed
the value $x^s_n$ itself, while the other half is allocated to embed
the unique identity for the node $X_n$. The value embedding is obtained by passing $x^s_n$, whether discrete or continuous, through an MLP\footnote{Using an MLP for a discrete variable is a slightly inefficient implementation of a node value embedding, but it ensures that the architecture is general.} encoder specific to node $X_n$. We use a node-specific embedding
because the values of each node may have very different interpretations and meanings. 
The node identity embedding is obtained using a standard 1D transformer positional embedding over node indices.
For column $S+1$ of the input, the value embedding is a vector of zeros.

\textbf{Alternating attention.} 
Traditional transformers discover relationships among the elements of a data sample arranged in a one-dimensional
sequence. With our two-dimensional lattice, the transformer could operate over the
entire lattice at once to discover relationships among both nodes and samples. 
Given an encoding that indicates position $n,s$ in the lattice, the model can in principle discover stable
relationships among nodes over samples. However, the inductive bias to encourage the 
model to leverage the lattice structure is weak. Additionally, the model is 
invariant to sample ordering, which is desirable because the samples are \textit{iid}. Therefore, we arrange our 
transformer in alternating layers. In the first layer of the pair,
attention operates across all nodes of a single sample $(x^s_1,\ldots,x^s_N)^\T$ to encode the
relationships among two or more nodes. In the second layer of the pair, 
attention operates across all samples for a given node  $(x^1_n,\ldots,x^S_n)$ to encode
information about the distribution of node values. Alternating attention in transformers was also done in \citet{kossen2021self}.

\textbf{Encoder summary.}
The encoder produces a \emph{summary} vector $e^{\text{sum}}_{n}$ with $H$ elements for each node $X_n$, which captures essential information about the node's behavior and its interactions with other nodes. The summary representation is formed independently for
each node and involves combining information across the $S$ samples (the columns of the lattice). 
This is achieved with a method often used with transformers that involves a weighted average based on how informative
each sample is. The weighting is obtained using the embeddings in column $S+1$ to form queries, and embeddings in
columns $1,\ldots,S$ to provide keys and values, and then using standard key-value attention.
\subsubsection{Decoder}
The decoder uses the summary information from the encoder to generate a prediction of the adjacency matrix $A$ of the underlying $\calG$. It operates sequentially, at each step producing a binary output 
indicating the prediction $\hat A_{k,l}$ of $A_{k,l}$, proceeding row by row. The decoder is an autoregressive transformer, meaning that each prediction $\hat A_{kl}$ is obtained based on all elements of $A$ previously predicted, as well as the summary produced by the encoder.
Our method does not enforce acyclicity. Although this could in principle yield cycles in the graph, in practice we observe strong performance regardless (Section \ref{sec:analysis}), likely due to the fact that training and evaluation graphs typically  studied (e.g., ER-1 and ER-2) are very sparse.
%in practice we observed strong performance regardless. 
Nevertheless, one could likely improve the results e.g. by using post-processing \citep{lippe2021efficient} or by extending the method with an accept-reject algorithm \citep{castelletti2022bcdag,alphacode}.

\textbf{Auxiliary loss.}
We found that autoregressive decoding of the flattened $N\times N$ adjacency matrix is too difficult for the decoder to learn alone. To provide additional inductive bias to facilitate learning of causal graphs, we added the auxiliary task of predicting the parents $A_{n,:}$ and children $A_{:,n}$ of node $X_n$ from the encoder summary, $e^{\text{sum}}_{n}$. This is achieved using an MLP to learn a mapping $f_n$, such that %$f_n(e^{\text{sum}}_{n})=(\pa(X_n),\ch(X_n)^{\T})$. 
$f_n(e^{\text{sum}}_{n})=(\hat A_{n,:},\hat A_{:,n}^{\T})$. 
While this prediction is redundant with the operation of the decoder, it short-circuits the autoregressive decoder and provides a strong
training signal to support proper training. % of the decoder.  
\section{Synthetic data}
In this section, we discuss identifiability and describe how synthetic data were generated for training and testing in Sections \ref{sec:in_distribution} and \ref{sec:ood_experiments}, and  training in Section \ref{sec:sim_to_real}.

\textbf{Identifiability.}
Dataset $\calD^i$ associated to CBN $\calM^i$ is given by $\calD^i=\{x^s\}_{s=1}^S$, where  $x^s:=(x^s_1,\ldots, x^s_N)^\T$ is either an observational or interventional data sample obtained by performing a hard intervention on a randomly selected node in $\calG^i$. 
As discussed in \cite{eberhardt2006n}, in the limit of an infinite amount of such single-node interventional data samples, $\calG^i$ is identifiable.
As our model defines a distribution over graphs, its predictions are meaningful even when the amount of data is insufficient for identifiability: in this case, the model would sample graphs that are compatible with the given data.
Empirically, we found that our model can make reasonable predictions even with a small amount of samples per intervention and improves as more samples are observed (Section ~\ref{sec:analysis}).

\textbf{Graph distribution.}\label{sec:StructureDistribution}
We specified a distribution over  $\calG$ in terms of the number of nodes $N$ (graph size) and number of edges (graph density) present in $\calG$. As shown in \citet{zheng2018dags,yu2019dag,ke2020dependency}, larger and denser graphs are more challenging to learn. We varied $N$ from $5$ to $80$. 
We used the Erdős–Rényi (ER) metric to vary density and evaluated our model on ER-1 and ER-2 graphs, as in \citet{yu2019dag,brouillard2020differentiable,scherrer2021learning}. We generated an adjacency matrix $A$ by first sampling a lower-triangular matrix to ensure that it represents a DAG, and by then permuting the order of the nodes to ensure random ordering.

\textbf{Conditional probability distributions.}\label{sec:CPDs}
We performed ancestral sampling on the underlying CBN.
We considered both continuous and discrete nodes. For continuous nodes, %with interventions, values are sampled from the uniform distribution $U[-1,1]$. For nodes without interventions,
we generated \emph{continuous data} using three different methods following similar setups in previous works: (i) linear models (\emph{linear data}) \citep{zheng2018dags,yu2019dag}, (ii) nonlinear with additive noise models (\emph{ANM data}) \citep{brouillard2020differentiable,lippe2021efficient}, and (iii) nonlinear with non-additive noise models using neural networks \emph{(NN data)} \citep{brouillard2020differentiable,lippe2021efficient}. For discrete nodes, 
we generated discrete data using two different methods: MLP (\emph{MLP data}) and Dirichlet (\emph{Dirichlet data}) conditional-probability table generators. %, which
%we refer to as  and , respectively.
%The MLP had two fully connected layers of hidden dimensionality $32$. 
Following past work
\citep{ke2020dependency,scherrer2021learning}, we used a randomly initialized network. 
The Dirichlet generator filled in the rows of a conditional probability table by sampling a categorical distribution from a Dirichlet prior with symmetric
parameters $\alpha$. 
(We remind the reader that this generative procedure is performed prior to node ordering being randomized for presentation to the learning model.)  For more details, please refer to Appendix \ref{appendix:synethtic_data}.
\vspace{-1mm}
\section{Related work}
\vspace{-1mm}
Methods for inferring causal graphs %from observational and interventional data 
can broadly be categorized into score-based (continuous optimization methods included), constraint-based, and asymmetry-based methods. Score-based methods search through the space of  possible candidate graphs, usually DAGs, and ranks them based on some scoring function  \citep{chickering2002optimal,cooper1999causal,goudet2017causal,hauser2012characterization,heckerman1995learning,tsamardinos2006max,huang2018generalized,zhu2019causal}. 
Recently, \citet{zheng2018dags,yu2019dag,lachapelle2019gradient} framed the structure search as a continuous optimization problem. % which can be seen as a way to optimize for the scoring function.
There also exist score-based methods that use a mix of continuous and discrete optimization  \citep{bengio2019meta,ke2020dependency,lippe2021efficient,scherrer2021learning}. Constraint-based methods \citep{monti2019causal,spirtes2000causation, sun2007kernel, zhang2012kernel,zhu2019causal} infer the DAG by analyzing conditional independencies in the data.
\citet{eatonuai} use dynamic programming techniques. %to accelerate Markov Chain Monte Carlo sampling in a Bayesian approach to 
%for structure learning for DAGs. 
Asymmetry-based methods %such as 
\citep{shimizu2006linear, hoyer2009nonlinear, Peters2011b, daniusis2012inferring, Budhathoki17, mitrovic2018causal} assume asymmetry between cause and effect in the data and  use this %information 
to estimate the causal structure.
\citet{peters2016causal, ghassami2017learning,rojas2018invariant,heinze2018causal}  exploit invariance across environments to infer causal structure. 
\citet{mooij2016joint} propose a modelling framework that leverages existing methods.  
\begin{wraptable}[12]{r}{0.6\textwidth}
\centering
\vspace{.5\baselineskip}
\begin{tabular}{l|c|c|c}
\toprule
{\bf \small Data Type} & {\small{\bf RCC} } & {\small {\bf DAG-EQ} } & {\bf \small CSIvA}  \\
\midrule
{\small Observational} & \checkmark & \checkmark & \checkmark   \\ 
{\small Interventional} & \text{\sffamily X} & \text{\sffamily X} & \checkmark  \\ 
\midrule
{\small Linear dependencies} & \checkmark & \checkmark & \checkmark   \\ 
{\small Non-linear dependencies} & \checkmark & \text{\sffamily X} & \checkmark   \\ 
\bottomrule
\end{tabular}
\caption{Data-type comparison between CSIvA and other supervised approaches to causal structure induction (RCC \citep{lopez2015randomized,lopez2015towards} and DAG-EQ \citep{li2020supervised}). 
}
\label{table:comparison}
\vspace{-1\baselineskip}
\end{wraptable}

Several learning-based methods have been proposed \citep{bengio2019meta,goudet2018learning,guyoncause,guyonchalearn, kalainathan2018sam,ke2020dependency,ke2020amortized,lachapelle2022disentanglement,lopez2015towards,wang2021ordering,zhu2019causal}. These works are mainly concerned with learning  only part of the causal induction pipeline, such as the scoring function, and hence are significantly different from our work, which uses an end-to-end supervised learning approach to learn to map from datasets to graphs.
Neural network methods equipped with learned masks exist in the literature \citep{douglas2017universal,alias2021neural,ivanov2018variational,li2019flow,yoon2018gain}, but only a few have been adapted to causal inference. Several transformer models \cite{goyal2022retrieval,kossen2021self,muller2021transformers} have been proposed for learning to map from datasets to targets. However, none 
have been applied to causal discovery, although  
\citet{lowe2022amortized} proposes a neural-network based approach for causal discovery on time-series data.
A few supervised learning approaches have been proposed either framing the task as a kernel mean embedding classification problem \citep{lopez2015randomized,lopez2015towards} or operating directly on covariance matrices \citep{li2020supervised} or binary classification problem of identifying v-structures \citep{dai2021ml4c}.  These models accept observational data only (see Table~\ref{table:comparison}), and because causal identifiability requires \emph{both observational and interventional} data, our model is in principle more powerful.
\vspace{-2mm}
\section{Experiments}
\vspace{-2mm}
We report on a series of experiments of increasing challenge to our supervised approach to causal structure induction. 
First, we examined whether CSIvA generalizes well on synthetic data for which the training and test distributions are identical 
(Section~\ref{sec:in_distribution}). This experiment tests whether the model can learn to map from a dataset to a structure.
Second, we examined generalization to an out-of-distribution (OOD) test distribution, and we determined hyperparameters of the synthetic data generating
process that are most robust to OOD testing (Section~\ref{sec:ood_experiments}). 
Third, we trained CSIvA using the hyperparameters from our second experiment and evaluated it on a different type of OOD test distribution from several naturalistic
CBNs (Section~\ref{sec:sim_to_real}). This  experiment is the most important test of our hypothesis that causal structure of synthetic datasets
can be a useful proxy for discovering causal structure in realistic settings. Lastly, we performed a set of ablation studies to analyze the performance of CSIvA under different settings (Section~\ref{sec:analysis}). Please refer to Appendix \ref{appendix:hyperparam} for details on the hyperparameters.

\textbf{Comparisons to baselines.} We compared CSIvA to a range of methods considered to be state-of-the-art in the literature,
ranging from classic to neural-network based causal discovery baselines. For both  in-distribution and OOD experiments, we compare to 4 very strong baselines: DAG-GNN \citep{yu2019dag}, non-linear ICP \citep{heinze2018invariant},  DCDI \citep{brouillard2020differentiable}, and ENCO \citep{lippe2021efficient}. For OOD experiments to naturalistic graphs, we compare to 5 additional baselines
\citep{chickering2002optimal,hauser2012characterization,zheng2018dags,gamella2020active,li2020supervised}. 
Because some methods use only observational data and others do not scale well to large N, we could not compare to all methods in all experiments. 
See Appendix \ref{appendix:baseline_choices} for further discussion of alternative methods and conditions under which they can be used.

\subsection{In-distribution experiments}
\label{sec:in_distribution}
\begin{figure}
     \centering
    \begin{subfigure}[t]{0.45\textwidth}
        \raisebox{-\height}{\includegraphics[width=\textwidth]{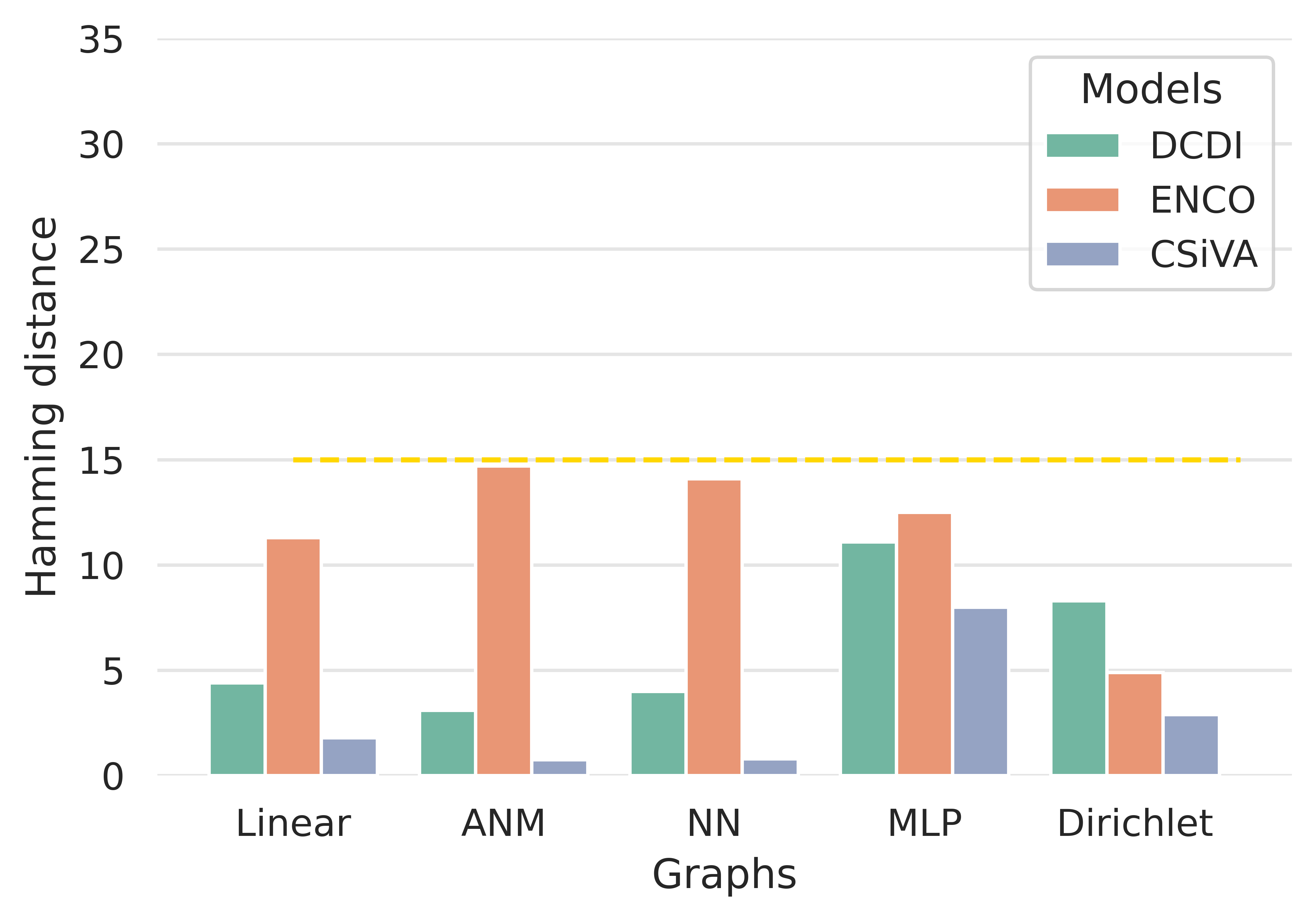}}
        \caption{Results on $N=30$ and ER-$1$ graphs.}
    \end{subfigure}
    % \hfill
    \begin{subfigure}[t]{0.45\textwidth}
        \raisebox{-\height}{\includegraphics[width=\textwidth]{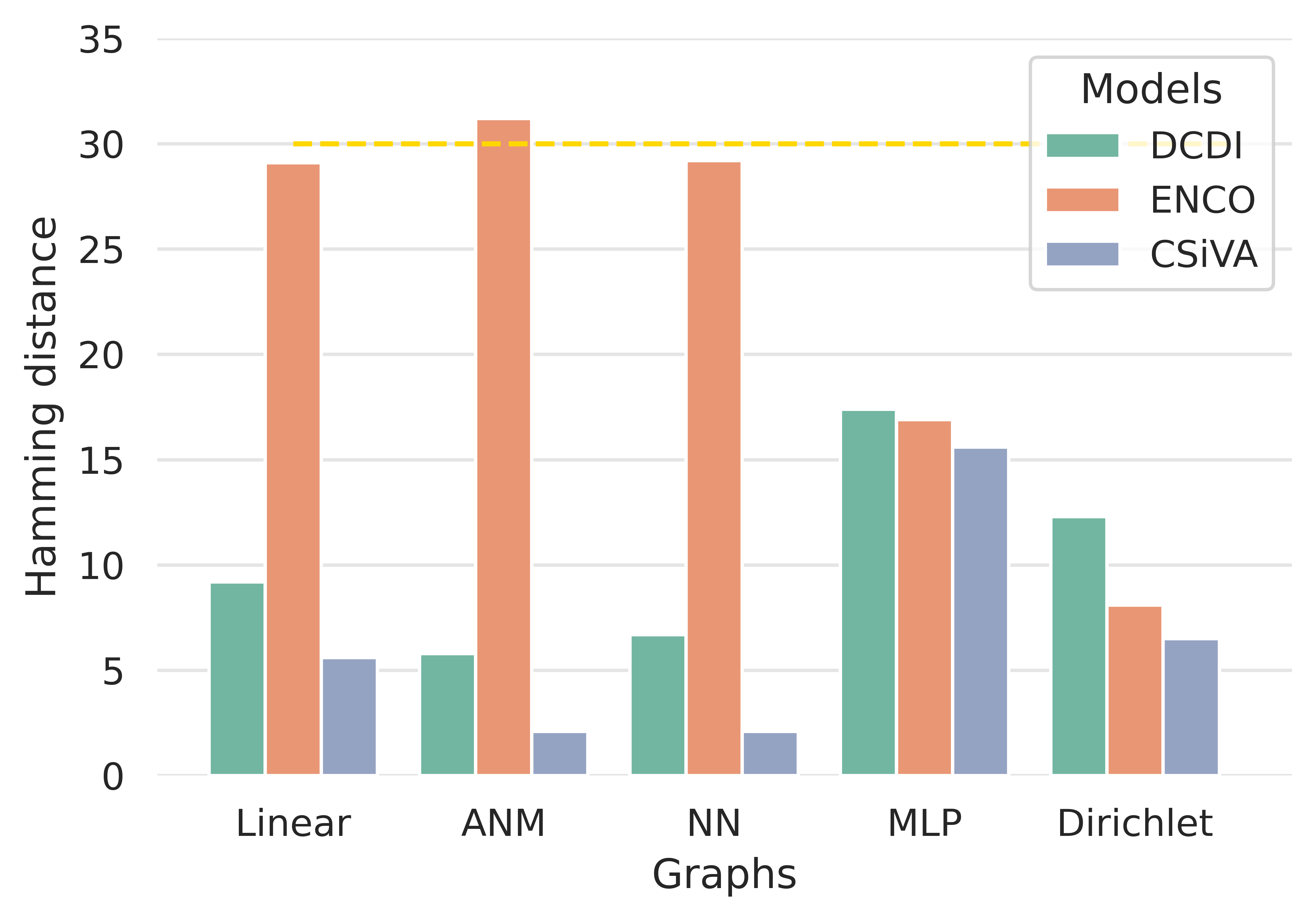}}
        \caption{Results on $N=30$ and ER-$2$ graphs.}
    \end{subfigure}
    %%%%%%%%%%%%%%%%%%%%%%%%%%%%%%%%%%%%second row
 \caption{\small Hamming distance \calH~between predicted and ground-truth adjacency matrices on 5 different data types: Linear, ANM, NN, MLP and Dirichlet data, compared to DCDI \citep{brouillard2020differentiable} and ENCO \citep{lippe2021efficient} on $N=30$ graphs, averaged over 128 sampled graphs. The dotted line indicates the value of the all-absent baseline. %Both non-linear ICP and \CSIVA~performs well on the easier (linear) continuous data. However, 
 \CSIVA~significantly outperforms all other baselines for all data types.}  
 \label{fig:N30_results}
\end{figure}

We begin by investigating whether \CSIVA~can learn to map from data to structures in the case in which the training and test distributions are identical. 
In this setting, our supervised approach has an advantage over unsupervised ones, as it can learn about the training distribution and leverage this knowledge during testing. We evaluate \CSIVA~ for $N\leq 80$ graphs.
We examined the performance on data with increasing order of difficulty, starting with linear (continuous data), before moving to non-linear cases (ANM, NN, MLP and Dirichlet data).  See Fig. \ref{fig:N30_results} for comparisons between  \CSIVA~and strong baselines on all data types ($N=30$), showing that \CSIVA~significantly outperforms  baselines for a wide range of data types.
For all but the Dirichlet data, we summarize the high-level results here, but detailed results and plots for all experiments can be found in Appendix \ref{appendix:in_distribution}.

\textbf{Linear data.} Results on continuous linear data for $N\leq20$ and $30\leq N \leq80$ are presented in Tables \ref{table:linear_results} and \ref{table:linear_larger_graphs_baselines} in the Appendix. \CSIVA~outperforms non-linear ICP, DAG-GNN, DCDI and ENCO. The difference in performance increases with $N$. %For details, please see Appendix \ref{appendix:linear_data}.

\textbf{ANM data.} Results on continuous ANM data for $N \leq80$ are presented in Table \ref{table:anm_larger_graphs_baselines} in the Appendix. \CSIVA~significantly outperforms the strongest baselines: DCDI and ENCO. Again, the difference in performance increases with $N$. %For more details, please see Appendix \ref{appendix:anm_data}.

\textbf{NN data.}
Results on continuous NN data are reported in Table \ref{table:nn_larger_graphs_baselines} in the Appendix. \CSIVA~significantly outperforms strongest baselines: DCDI and ENCO, with improved benefit for larger $N$. 

\textbf{MLP data.} Results on MLP data are shown in \figref{fig:continous_mlp_results}(b) in the Appendix.
\CSIVA~ significantly outperforms non-linear ICP, DAG-GNN, DCDI and ENCO. Differences become more apparent with larger graph sizes ($N\geq10$) and denser graphs (ER-2 vs ER-1).  

\textbf{Dirichlet data.} The Dirichlet data requires setting the values of the parameter $\alpha$. Hence, we run two sets of experiments on this data.  In the first set of experiments, we investigated how different values of $\alpha$ impact learning in \CSIVA. As shown in Table \ref{table:dirichlet_data_all_alpha} in the appendix, \CSIVA~performs well on all data with $\alpha \leq 0.5$, achieving $\mathcal H<2.5$ in all cases. \CSIVA~still performs well when $\alpha=1.0$, achieving $\mathcal H<5$ on size $10$ graphs. Learning with $\alpha>1$ is more challenging. This is not surprising, as $\alpha>1$ tends to generate more uniform distributions, which are not informative of the causal relationship between nodes.

\begin{wrapfigure}[24]{r}{0.6\textwidth}
    \vspace{-0.8\baselineskip}
    \includegraphics[width=0.6\textwidth]{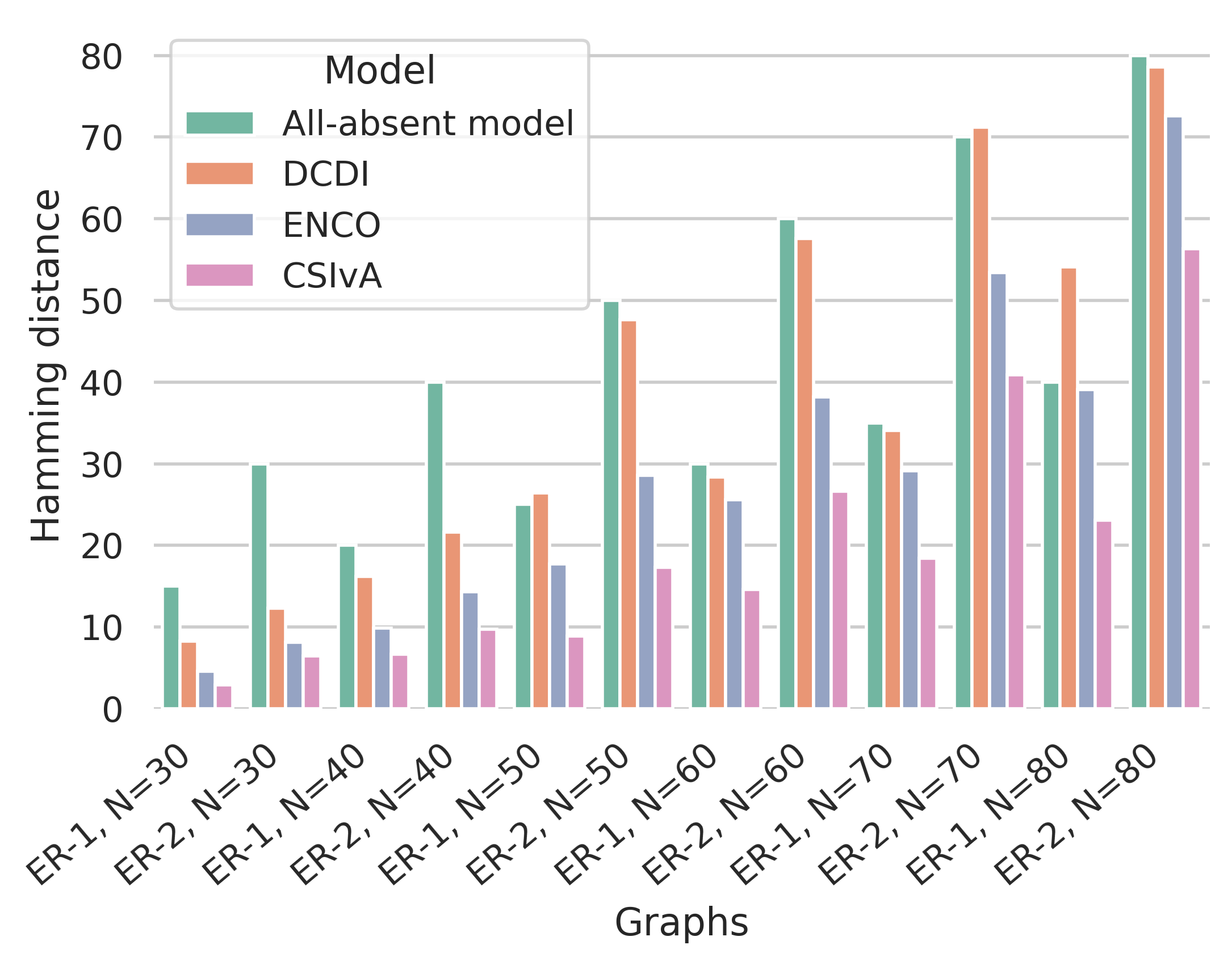}
    \caption{\textbf{Results on Dirichlet data for $30\leq N \leq80$}. Hamming distance \calH~~between predicted and ground-truth adjacency matrices, averaged over 128 sampled graphs. \CSIVA~ significantly outperforms DCDI and ENCO, both of which are very strong baselines. The difference in performance increases with $N$.}
    % \vspace{-.5\baselineskip}
    \label{fig:dirichlet_results_large}
\end{wrapfigure}

In the second set of experiments, we compared \CSIVA~to DCDI and ENCO, as they are the strongest performing baselines. We run the comparisons on graphs of size up to $N\leq80$. To limit the number of experiments to run, we set $\alpha=0.1$, as this allows the conditional probabilities to concentrate on non-uniform distributions, which is more likely to be true in realistic settings. As shown in  \figref{fig:dirichlet_results_large}, our model significantly outperforms DCDI and ENCO on all graphs (up to $N\leq80$). As with other data types, the difference becomes more apparent as the size of the graphs grows ($N\geq50$), as  larger graphs are more challenging to learn.

\textbf{Soft interventions and unknown interventions.} We additionally evaluated \CSIVA~on soft interventions, and on hard interventions for which the model did not know the intervention node (\emph{unknown interventions}). In both cases, we focus on Dirichlet data with $\alpha=0.1$  and $\alpha=0.25$ for ER-1 and ER-2 graphs of size $N\leq20$.
Results for the performance of \CSIVA~on soft interventions are shown in Table \ref{table:imperfect_interventions} in the Appendix. \CSIVA~ performed well on all graphs, achieving hamming distance $\mathcal H<5$ on all graphs. For details about the data generation process, refer to Appendix Section \ref{sec:imperfect_interventions}.

Results for the performance of \CSIVA~on unknown interventions  are shown in Appendix Table \ref{table:unknown_interventions}. \CSIVA~achieves strong performance on all graphs, with hamming distance $\mathcal H<7$ on all graphs. Note that \CSIVA~loses almost no performance despite not knowing the intervention; the biggest difference in hamming distance  between known and unknown interventions on all graphs is $\mathcal H<1.0$. For more details, refer to Section \ref{sec:appendix_unknown_intervention} in the Appendix.

\subsection{Out-of-distribution experiments}\label{sec:ood_experiments}
In this set of experiments, we evaluated \CSIVA's ability to generalize with respect to aspects of the data generating distribution that are often unknown, namely graph density and parameters of the \cpds, such as the $\alpha$ values of the Dirichlet distribution. 

\textbf{Varying graph density.} 
We evaluated how well our model performs when trained and tested on CBNs with varying graph density on MLP and $\alpha=1$ Dirichlet data. 
We fixed the number of nodes to $N=7$, with variables able to take on discrete values in $\{1, 2, 3 \}$. The graphs in training and test datasets 
can take ER degree $\in \{1, 2, 3 \}$. 
Results are shown in Table \ref{table:ood_generalization_mlp} in the appendix for the MLP data and Table \ref{table:ood_generalization_ER_dirichlet} in the appendix for the Dirichlet data. For the MLP data, models trained on ER-$2$ graph generalizes the best. For Dirichlet data, there isn't one value of graph density that consistently generalizes best across graphs with different densities. Nevertheless, ER-$2$ graphs give a balanced trade-off and generalizes well across graphs with different sparsity.

\textbf{Varying $\alpha$.} We next trained and evaluated on data generated from Dirichlet distributions with 
$\alpha \in \{0.1, 0.25, 0.5\}$. Results for ER-$1$ graphs with $N=7$ are found in Table \ref{table:ood_generalization_alpha}. 
There isn't a value of $\alpha$ that performs consistently well across  different values of $\alpha$ for the test data. Nevertheless, $\alpha=0.25$ is a balanced trade-off and generalizes well across test data with $0.1 \leq \alpha \leq 0.5$.
\vspace{-1mm}
\subsection{Sim-to-real experiments}\label{sec:sim_to_real}
\vspace{-1mm}

In this final set of experiments, we evaluated \CSIVA's ability to generalize from being trained on MLP and Dirichlet data to being evaluated on the widely used Sachs \citep{sachs2005causal}, 
Asia \citep{lauritzen1988local} and Child \citep{Spiegelhalter1992Learning} CBNs from the BnLearn repository, which have $N=11$, $N=8$ and $N=20$ nodes respectively. We 
followed the established protocol from \citet{ke2020dependency,lippe2021efficient,scherrer2021learning}
of sampling observational and interventional data from the CBNs provided by the repository. These experiments are the most important test of our hypothesis that causal structure of synthetic datasets
can be a useful proxy for discovering causal structure in realistic settings.

We emphasize that all hyperparameters for the MLP and Dirichlet data and for the learning procedure were chosen \textit{a priori}; only after the architecture and parameters were finalized was 
the model tested on these benchmarks. 
Furthermore, to keep the setup simple, 
we trained on data sampled from a single set of hyperparameters instead of a broader mixture. 
Findings in Section \ref{sec:ood_experiments} suggest that ER-$2$ graphs with $\alpha=0.25$ 
work well overall and hence were chosen. 
Results are reported in Table \ref{table:real_baseline_hamming}, comparing to a range of baselines from  \citet{squires2020permutation,heinze2018invariant,yu2019dag,gamella2020active,brouillard2020differentiable,lippe2021efficient} and others.

\begin{wraptable}[20]{r}{9.5cm}
\centering  
{\small
\begin{tabular}{lrrr}
\toprule
& {\bf \small  Asia} &{\bf \small  Sachs} & {\bf \small Child} \\
\midrule
{\small Number of nodes} & 8 & 11 & 20 \\
%\midrule
{\small All-absent Baseline}  & 8 & 17 & 25 \\
\midrule
%All-present Baseline & -  &  &  \\
{\small GES \cite{chickering2002optimal}}  & 4 & 19  & 33\\ 
{\small DAG-notears \cite{zheng2018dags}} & 14 & 22    & 23   \\
{\small DAG-GNN \cite{yu2019dag}}    & 8 &  13 & 20\\
\midrule
{\small GES \cite{hauser2012characterization}}  & 11 &  16  & 31 \\ 
%{\bf \small \citet{gamella2020active}}   &  17 & 17        \\
{\small ICP \cite{peters2016causal}}   & 8   &  17 &  $27^{\small{*}}$  \\
{\small Non-linear ICP \cite{heinze2018invariant}} & 8  &  16 & $23^{\small{*}}$ \\
\midrule
{\small DAG-EQ \cite{li2020supervised}}  & - & 16  & - \\
%{\small IGSP \cite{squires2020permutation}}  & 18 & - & - \\
%{\small DCDI-G \cite{brouillard2020differentiable}}  & 36 & - & - \\
{\small DCDI-DSF \cite{brouillard2020differentiable}} & 7 & 33  & 18 
\\
{\small ENCO \cite{lippe2021efficient}}   & 5 & 9 & 14 
\\
\midrule
CSIvA~(MLP data) &  \colorbox{lightmintbg}{\textbf{3}}  &
\colorbox{lightmintbg}{\textbf{6}} & \colorbox{lightmintbg}{\textbf{11}}\\ 
CSIvA~ (Dirichlet data) &   \colorbox{lightmintbg}{\textbf{3}}  & \colorbox{lightmintbg}{\textbf{5}} & \colorbox{lightmintbg}{\textbf{10}}\\
\bottomrule
\end{tabular}
}
\caption{\textbf{Results on Asia, Sachs and 
Child data:} Hamming distance \calH~between predicted and ground-truth adjacency matrices.{ \small{*}}To maintain computational tractability, the number of parents was limited to 3.  %\CSIVA~significantly outperforms other models.
}
%\vspace{-.2\baselineskip}
\label{table:real_baseline_hamming}
\end{wraptable}
% Note that we do not compare to the method in \citet{ke2020dependency}, as this method needs at least $500,000$ data samples (which is more than $300$ times the amount required by our method).
\CSIVA~trained on both MLP and  Dirichlet data significantly outperforms all other methods on both the Asia and Sachs datasets. This serves as strong evidence  that our model can learn to induce causal structures in the more realistic real-world CBNs, while only trained on synthetic data.

\vspace{-2mm}
\subsection{Detailed analyses}

\label{sec:analysis}
In order to better understand model performance, and which model components were most critical, we performed further analyses. We describe here a subset of the studies performed; please refer to Appendix \ref{sec:appendix_ablations} for the full results, including ablations. 

\textbf{Acyclicity.} We analyzed the generated (sampled) graphs for acyclicity. We evaluated using Dirichlet data with $\alpha=0.1$ and $\alpha=0.25$ on ER-1 and ER-2 graphs of size $N\leq20$. For each setting, we evaluated the proposed model on $128$ random graphs, and we found that none of the generated graphs contains any cycles, i.e., all generated graphs are acyclic.

\textbf{Visualization of generated graphs.} We visualized some of the generated graphs in Figure \ref{fig:samples_er1} and Figure \ref{fig:samples_er2} in the Appendix. Note that samples are randomly drawn, and all are acyclic.

\textbf{Identifiability upon seeing intervention data.} Here, we investigate how varying the proportion of interventional data impacts the performance of the proposed model. As shown in Figure \ref{fig:percent_intervention} (Appendix), the model's performance improves (Hamming distance decreases) almost monotonically as the amount of interventional data increases, from $0-100\%$. This is a clear indication that our model is able to extract information from interventional data for identifying the graph structure. For more details on the results and the how the experiments are conducted, please refer to Section \ref{sec:appendix_percent_interventions}.

\textbf{Ablation studies.}
We investigate the role of various components in  \CSIVA. For example, what is the effect of having the auxiliary loss or the sample-level attention. We found that all the different components play an important role in our model. For details, see Appendix \ref{sec:appendix_ablations}.

\section{Discussion}
In this paper, we have presented a novel approach towards causal graph structure inference. %using observational and interventional data. 
Our method is based on learning from synthetic data in order to obtain a strong learning signal (in the form of explicit supervision), using a novel transformer-based architecture which directly analyzes the data and computes a distribution of candidate graphs. Through a comprehensive and detailed set of experiments, we demonstrated that even though only trained on synthetic data, our model generalizes to out-of-distribution datasets, and robustly outperforms comparison methods under a wide range of conditions. 
A direction of future work would be to use the proposed framework for learning causal structure from raw visual data. This could be useful, e.g. in an RL setting in which an RL agent interacts with the environment via observing low level pixel data \citep{ahmed2020causalworld, ke2021systematic,wang2021alchemy}.
\bibliography{iclr2023_conference}

\begin{thebibliography}{75}
\providecommand{\natexlab}[1]{#1}
\providecommand{\url}[1]{\texttt{#1}}
\expandafter\ifx\csname urlstyle\endcsname\relax
  \providecommand{\doi}[1]{doi: #1}\else
  \providecommand{\doi}{doi: \begingroup \urlstyle{rm}\Url}\fi

\bibitem[Ahmed et~al.(2020)Ahmed, Tr{\"a}uble, Goyal, Neitz, Bengio,
  Sch{\"o}lkopf, W{\"u}thrich, and Bauer]{ahmed2020causalworld}
Ossama Ahmed, Frederik Tr{\"a}uble, Anirudh Goyal, Alexander Neitz, Yoshua
  Bengio, Bernhard Sch{\"o}lkopf, Manuel W{\"u}thrich, and Stefan Bauer.
\newblock Causalworld: A robotic manipulation benchmark for causal structure
  and transfer learning.
\newblock \emph{arXiv preprint arXiv:2010.04296}, 2020.

\bibitem[Ba et~al.(2016)Ba, Kiros, and Hinton]{ba2016layer}
Jimmy~Lei Ba, Jamie~Ryan Kiros, and Geoffrey~E Hinton.
\newblock Layer normalization.
\newblock \emph{arXiv preprint arXiv:1607.06450}, 2016.

\bibitem[Bengio et~al.(2019)Bengio, Deleu, Rahaman, Ke, Lachapelle, Bilaniuk,
  Goyal, and Pal]{bengio2019meta}
Yoshua Bengio, Tristan Deleu, Nasim Rahaman, Rosemary Ke, S{\'e}bastien
  Lachapelle, Olexa Bilaniuk, Anirudh Goyal, and Christopher Pal.
\newblock A meta-transfer objective for learning to disentangle causal
  mechanisms.
\newblock \emph{arXiv preprint arXiv:1901.10912}, 2019.

\bibitem[Brouillard et~al.(2020)Brouillard, Lachapelle, Lacoste,
  Lacoste-Julien, and Drouin]{brouillard2020differentiable}
Philippe Brouillard, S{\'e}bastien Lachapelle, Alexandre Lacoste, Simon
  Lacoste-Julien, and Alexandre Drouin.
\newblock Differentiable causal discovery from interventional data.
\newblock \emph{Advances in Neural Information Processing Systems},
  33:\penalty0 21865--21877, 2020.

\bibitem[Budhathoki \& Vreeken(2017)Budhathoki and Vreeken]{Budhathoki17}
Kailash Budhathoki and Jilles Vreeken.
\newblock Causal inference by stochastic complexity.
\newblock \emph{arXiv:1702.06776}, 2017.

\bibitem[Castelletti \& Mascaro(2022)Castelletti and
  Mascaro]{castelletti2022bcdag}
Federico Castelletti and Alessandro Mascaro.
\newblock Bcdag: An r package for bayesian structure and causal learning of
  gaussian dags.
\newblock \emph{arXiv preprint arXiv:2201.12003}, 2022.

\bibitem[Chickering(2002)]{chickering2002optimal}
David~Maxwell Chickering.
\newblock Optimal structure identification with greedy search.
\newblock \emph{Journal of machine learning research}, 3\penalty0
  (Nov):\penalty0 507--554, 2002.

\bibitem[Cooper \& Yoo(1999)Cooper and Yoo]{cooper1999causal}
Gregory~F. Cooper and Changwon Yoo.
\newblock Causal {Discovery} from a {Mixture} of {Experimental} and
  {Observational} {Data}.
\newblock In \emph{Proceedings of the {Fifteenth} {Conference} on {Uncertainty}
  in {Artificial} {Intelligence}}, {UAI}'99, pp.\  116--125, San Francisco, CA,
  USA, 1999.

\bibitem[Dai et~al.(2021)Dai, Ding, Jiang, Han, and Zhang]{dai2021ml4c}
Haoyue Dai, Rui Ding, Yuanyuan Jiang, Shi Han, and Dongmei Zhang.
\newblock Ml4c: Seeing causality through latent vicinity.
\newblock \emph{arXiv preprint arXiv:2110.00637}, 2021.

\bibitem[Daniusis et~al.(2012)Daniusis, Janzing, Mooij, Zscheischler, Steudel,
  Zhang, and Sch{\"o}lkopf]{daniusis2012inferring}
Povilas Daniusis, Dominik Janzing, Joris Mooij, Jakob Zscheischler, Bastian
  Steudel, Kun Zhang, and Bernhard Sch{\"o}lkopf.
\newblock Inferring deterministic causal relations.
\newblock \emph{arXiv preprint arXiv:1203.3475}, 2012.

\bibitem[Devlin et~al.(2018)Devlin, Chang, Lee, and Toutanova]{devlin2018bert}
Jacob Devlin, Ming-Wei Chang, Kenton Lee, and Kristina Toutanova.
\newblock Bert: Pre-training of deep bidirectional transformers for language
  understanding.
\newblock \emph{arXiv preprint arXiv:1810.04805}, 2018.

\bibitem[Douglas et~al.(2017)Douglas, Zarov, Gourgoulias, Lucas, Hart, Baker,
  Sahani, Perov, and Johri]{douglas2017universal}
Laura Douglas, Iliyan Zarov, Konstantinos Gourgoulias, Chris Lucas, Chris Hart,
  Adam Baker, Maneesh Sahani, Yura Perov, and Saurabh Johri.
\newblock A universal marginalizer for amortized inference in generative
  models.
\newblock \emph{arXiv preprint arXiv:1711.00695}, 2017.

\bibitem[Drton \& Maathuis(2017)Drton and Maathuis]{drton2017structure}
Mathias Drton and Marloes~H. Maathuis.
\newblock Structure learning in graphical modeling.
\newblock \emph{Annual Review of Statistics and Its Application}, 4\penalty0
  (1):\penalty0 365--393, 2017.

\bibitem[Eaton \& Murphy(2007)Eaton and Murphy]{eatonuai}
Daniel Eaton and Kevin Murphy.
\newblock {Bayesian structure learning using dynamic programming and MCMC}.
\newblock In \emph{Uncertainty in Artificial Intelligence}, pp.\  101--108,
  2007.

\bibitem[Eberhardt et~al.(2006)Eberhardt, Glymour, and
  Scheines]{eberhardt2006n}
Frederick Eberhardt, Clark Glymour, and Richard Scheines.
\newblock N-1 experiments suffice to determine the causal relations among n
  variables.
\newblock In \emph{Innovations in machine learning}, pp.\  97--112. Springer,
  2006.

\bibitem[Gamella \& Heinze-Deml(2020)Gamella and
  Heinze-Deml]{gamella2020active}
Juan~L Gamella and Christina Heinze-Deml.
\newblock Active invariant causal prediction: Experiment selection through
  stability.
\newblock \emph{arXiv preprint arXiv:2006.05690}, 2020.

\bibitem[Ghassami et~al.(2017)Ghassami, Salehkaleybar, Kiyavash, and
  Zhang]{ghassami2017learning}
AmirEmad Ghassami, Saber Salehkaleybar, Negar Kiyavash, and Kun Zhang.
\newblock Learning causal structures using regression invariance.
\newblock In \emph{Advances in Neural Information Processing Systems}, pp.\
  3011--3021, 2017.

\bibitem[Glymour et~al.(2019)Glymour, Zhang, and Spirtes]{glymor2019review}
Clark Glymour, Kun Zhang, and Peter Spirtes.
\newblock Review of causal discovery methods based on graphical models.
\newblock \emph{Frontiers in Genetics}, 10:\penalty0 524, 2019.

\bibitem[Goudet et~al.(2017)Goudet, Kalainathan, Caillou, Guyon, Lopez-Paz, and
  Sebag]{goudet2017causal}
Olivier Goudet, Diviyan Kalainathan, Philippe Caillou, Isabelle Guyon, David
  Lopez-Paz, and Mich{\`e}le Sebag.
\newblock Causal generative neural networks.
\newblock \emph{arXiv preprint arXiv:1711.08936}, 2017.

\bibitem[Goudet et~al.(2018)Goudet, Kalainathan, Caillou, Guyon, Lopez-Paz, and
  Sebag]{goudet2018learning}
Olivier Goudet, Diviyan Kalainathan, Philippe Caillou, Isabelle Guyon, David
  Lopez-Paz, and Michele Sebag.
\newblock Learning functional causal models with generative neural networks.
\newblock In \emph{Explainable and Interpretable Models in Computer Vision and
  Machine Learning}, pp.\  39--80. Springer, 2018.

\bibitem[Goyal et~al.(2021)Goyal, Didolkar, Ke, Blundell, Beaudoin, Heess,
  Mozer, and Bengio]{alias2021neural}
Anirudh Goyal, Aniket Didolkar, Nan~Rosemary Ke, Charles Blundell, Philippe
  Beaudoin, Nicolas Heess, Michael~C Mozer, and Yoshua Bengio.
\newblock Neural production systems.
\newblock \emph{Advances in Neural Information Processing Systems}, 34, 2021.

\bibitem[Goyal et~al.(2022)Goyal, Friesen, Banino, Weber, Ke, Badia, Guez,
  Mirza, Konyushkova, Valko, et~al.]{goyal2022retrieval}
Anirudh Goyal, Abram~L Friesen, Andrea Banino, Theophane Weber, Nan~Rosemary
  Ke, Adria~Puigdomenech Badia, Arthur Guez, Mehdi Mirza, Ksenia Konyushkova,
  Michal Valko, et~al.
\newblock Retrieval-augmented reinforcement learning.
\newblock \emph{arXiv preprint arXiv:2202.08417}, 2022.

\bibitem[Guyon(2013)]{guyoncause}
Isabelle Guyon.
\newblock Cause-effect pairs kaggle competition, 2013.
\newblock \emph{URL https://www. kaggle. com/c/cause-effect-pairs}, pp.\  165,
  2013.

\bibitem[Guyon(2014)]{guyonchalearn}
Isabelle Guyon.
\newblock Chalearn fast causation coefficient challenge, 2014.
\newblock \emph{URL https://www. codalab. org/competitions/1381}, pp.\  165,
  2014.

\bibitem[Hauser \& B{\"u}hlmann(2012)Hauser and
  B{\"u}hlmann]{hauser2012characterization}
Alain Hauser and Peter B{\"u}hlmann.
\newblock Characterization and greedy learning of interventional markov
  equivalence classes of directed acyclic graphs.
\newblock \emph{The Journal of Machine Learning Research}, 13\penalty0
  (1):\penalty0 2409--2464, 2012.

\bibitem[He et~al.(2016)He, Zhang, Ren, and Sun]{he2016deep}
Kaiming He, Xiangyu Zhang, Shaoqing Ren, and Jian Sun.
\newblock Deep residual learning for image recognition.
\newblock In \emph{Proceedings of the IEEE conference on computer vision and
  pattern recognition}, pp.\  770--778, 2016.

\bibitem[Heckerman et~al.(1995)Heckerman, Geiger, and
  Chickering]{heckerman1995learning}
David Heckerman, Dan Geiger, and David~M Chickering.
\newblock Learning bayesian networks: The combination of knowledge and
  statistical data.
\newblock \emph{Machine learning}, 20\penalty0 (3):\penalty0 197--243, 1995.

\bibitem[Heinze-Deml et~al.(2018{\natexlab{a}})Heinze-Deml, Maathuis, and
  Meinshausen]{heinze2018causal}
Christina Heinze-Deml, Marloes~H Maathuis, and Nicolai Meinshausen.
\newblock Causal structure learning.
\newblock \emph{Annual Review of Statistics and Its Application}, 5:\penalty0
  371--391, 2018{\natexlab{a}}.

\bibitem[Heinze-Deml et~al.(2018{\natexlab{b}})Heinze-Deml, Peters, and
  Meinshausen]{heinze2018invariant}
Christina Heinze-Deml, Jonas Peters, and Nicolai Meinshausen.
\newblock Invariant causal prediction for nonlinear models.
\newblock \emph{Journal of Causal Inference}, 6\penalty0 (2),
  2018{\natexlab{b}}.

\bibitem[Hoyer et~al.(2009)Hoyer, Janzing, Mooij, Peters, and
  Sch{\"o}lkopf]{hoyer2009nonlinear}
Patrik~O Hoyer, Dominik Janzing, Joris~M Mooij, Jonas Peters, and Bernhard
  Sch{\"o}lkopf.
\newblock Nonlinear causal discovery with additive noise models.
\newblock In \emph{Advances in neural information processing systems}, pp.\
  689--696, 2009.

\bibitem[Huang et~al.(2018)Huang, Zhang, Lin, Sch{\"o}lkopf, and
  Glymour]{huang2018generalized}
Biwei Huang, Kun Zhang, Yizhu Lin, Bernhard Sch{\"o}lkopf, and Clark Glymour.
\newblock Generalized score functions for causal discovery.
\newblock In \emph{Proceedings of the 24th ACM SIGKDD international conference
  on knowledge discovery \& data mining}, pp.\  1551--1560, 2018.

\bibitem[Ivanov et~al.(2018)Ivanov, Figurnov, and
  Vetrov]{ivanov2018variational}
Oleg Ivanov, Michael Figurnov, and Dmitry Vetrov.
\newblock Variational autoencoder with arbitrary conditioning.
\newblock \emph{arXiv preprint arXiv:1806.02382}, 2018.

\bibitem[Kalainathan et~al.(2018)Kalainathan, Goudet, Guyon, Lopez-Paz, and
  Sebag]{kalainathan2018sam}
Diviyan Kalainathan, Olivier Goudet, Isabelle Guyon, David Lopez-Paz, and
  Mich{\`e}le Sebag.
\newblock Sam: Structural agnostic model, causal discovery and penalized
  adversarial learning.
\newblock \emph{arXiv preprint arXiv:1803.04929}, 2018.

\bibitem[Ke et~al.(2020{\natexlab{a}})Ke, Bilaniuk, Goyal, Bauer,
  Sch{\"o}lkopf, Mozer, Larochelle, Pal, and Bengio]{ke2020dependency}
Nan~Rosemary Ke, Olexa Bilaniuk, Anirudh Goyal, Stefan Bauer, Bernhard
  Sch{\"o}lkopf, Michael~Curtis Mozer, Hugo Larochelle, Christopher Pal, and
  Yoshua Bengio.
\newblock Dependency structure discovery from interventions.
\newblock 2020{\natexlab{a}}.

\bibitem[Ke et~al.(2020{\natexlab{b}})Ke, Wang, Mitrovic, Szummer, Rezende,
  et~al.]{ke2020amortized}
Nan~Rosemary Ke, Jane Wang, Jovana Mitrovic, Martin Szummer, Danilo~J Rezende,
  et~al.
\newblock Amortized learning of neural causal representations.
\newblock \emph{arXiv preprint arXiv:2008.09301}, 2020{\natexlab{b}}.

\bibitem[Ke et~al.(2021)Ke, Didolkar, Mittal, Goyal, Lajoie, Bauer, Rezende,
  Mozer, Bengio, and Pal]{ke2021systematic}
Nan~Rosemary Ke, Aniket~Rajiv Didolkar, Sarthak Mittal, Anirudh Goyal,
  Guillaume Lajoie, Stefan Bauer, Danilo~Jimenez Rezende, Michael~Curtis Mozer,
  Yoshua Bengio, and Christopher Pal.
\newblock Systematic evaluation of causal discovery in visual model based
  reinforcement learning.
\newblock 2021.

\bibitem[Kingma \& Ba(2014)Kingma and Ba]{kingma2014adam}
Diederik~P Kingma and Jimmy Ba.
\newblock Adam: A method for stochastic optimization.
\newblock \emph{arXiv preprint arXiv:1412.6980}, 2014.

\bibitem[Koller \& Friedman(2009)Koller and Friedman]{kollerl2009probabilistic}
Daphne Koller and Nir Friedman.
\newblock \emph{Probabilistic Graphical Models: Principles and Techniques}.
\newblock MIT Press, 2009.

\bibitem[Koski \& Noble(2012)Koski and Noble]{koski2012review}
Timo Koski and John Noble.
\newblock A review of {B}ayesian networks and structure learning.
\newblock \emph{Mathematica Applicanda}, 40, 2012.

\bibitem[Kossen et~al.(2021)Kossen, Band, Lyle, Gomez, Rainforth, and
  Gal]{kossen2021self}
Jannik Kossen, Neil Band, Clare Lyle, Aidan~N Gomez, Thomas Rainforth, and
  Yarin Gal.
\newblock Self-attention between datapoints: Going beyond individual
  input-output pairs in deep learning.
\newblock \emph{Advances in Neural Information Processing Systems}, 34, 2021.

\bibitem[Lachapelle et~al.(2019)Lachapelle, Brouillard, Deleu, and
  Lacoste-Julien]{lachapelle2019gradient}
S{\'e}bastien Lachapelle, Philippe Brouillard, Tristan Deleu, and Simon
  Lacoste-Julien.
\newblock Gradient-based neural dag learning.
\newblock \emph{arXiv preprint arXiv:1906.02226}, 2019.

\bibitem[Lachapelle et~al.(2022)Lachapelle, Rodriguez, Sharma, Everett,
  Le~Priol, Lacoste, and Lacoste-Julien]{lachapelle2022disentanglement}
S{\'e}bastien Lachapelle, Pau Rodriguez, Yash Sharma, Katie~E Everett, R{\'e}mi
  Le~Priol, Alexandre Lacoste, and Simon Lacoste-Julien.
\newblock Disentanglement via mechanism sparsity regularization: A new
  principle for nonlinear ica.
\newblock In \emph{Conference on Causal Learning and Reasoning}, pp.\
  428--484. PMLR, 2022.

\bibitem[Lauritzen \& Spiegelhalter(1988)Lauritzen and
  Spiegelhalter]{lauritzen1988local}
Steffen~L Lauritzen and David~J Spiegelhalter.
\newblock Local computations with probabilities on graphical structures and
  their application to expert systems.
\newblock \emph{Journal of the Royal Statistical Society: Series B
  (Methodological)}, 50\penalty0 (2):\penalty0 157--194, 1988.

\bibitem[Li et~al.(2020)Li, Xiao, and Tian]{li2020supervised}
Hebi Li, Qi~Xiao, and Jin Tian.
\newblock Supervised whole dag causal discovery.
\newblock \emph{arXiv preprint arXiv:2006.04697}, 2020.

\bibitem[Li et~al.(2019)Li, Akbar, and Oliva]{li2019flow}
Yang Li, Shoaib Akbar, and Junier~B Oliva.
\newblock Flow models for arbitrary conditional likelihoods.
\newblock \emph{arXiv preprint arXiv=1909.06319}, 2019.

\bibitem[Li et~al.(2022)Li, Choi, Chung, Kushman, Schrittwieser, Leblond,
  Eccles, Keeling, Gimeno, Lago, Hubert, Choy, d'Autume, Babuschkin, Chen,
  Huang, Welbl, Gowal, Cherepanov, Molloy, Mankowitz, Robson, Kohli,
  de~Freitas, Kavukcuoglu, and Vinyals]{alphacode}
Yujia Li, David Choi, Junyoung Chung, Nate Kushman, Julian Schrittwieser, Rémi
  Leblond, Tom Eccles, James Keeling, Felix Gimeno, Agustin~Dal Lago, Thomas
  Hubert, Peter Choy, Cyprien de~Masson d'Autume, Igor Babuschkin, Xinyun Chen,
  Po-Sen Huang, Johannes Welbl, Sven Gowal, Alexey Cherepanov, James Molloy,
  Daniel~J. Mankowitz, Esme~Sutherland Robson, Pushmeet Kohli, Nando
  de~Freitas, Koray Kavukcuoglu, and Oriol Vinyals.
\newblock Competition-level code generation with alphacode.
\newblock \emph{arXiv preprint arXiv:2203.07814}, 2022.

\bibitem[Lippe et~al.(2021)Lippe, Cohen, and Gavves]{lippe2021efficient}
Phillip Lippe, Taco Cohen, and Efstratios Gavves.
\newblock Efficient neural causal discovery without acyclicity constraints.
\newblock \emph{arXiv preprint arXiv:2107.10483}, 2021.

\bibitem[Lopez-Paz et~al.(2015{\natexlab{a}})Lopez-Paz, Muandet, and
  Recht]{lopez2015randomized}
David Lopez-Paz, Krikamol Muandet, and Benjamin Recht.
\newblock The randomized causation coefficient.
\newblock \emph{J. Mach. Learn. Res.}, 16:\penalty0 2901--2907,
  2015{\natexlab{a}}.

\bibitem[Lopez-Paz et~al.(2015{\natexlab{b}})Lopez-Paz, Muandet, Sch{\"o}lkopf,
  and Tolstikhin]{lopez2015towards}
David Lopez-Paz, Krikamol Muandet, Bernhard Sch{\"o}lkopf, and Iliya
  Tolstikhin.
\newblock Towards a learning theory of cause-effect inference.
\newblock In \emph{International Conference on Machine Learning}, pp.\
  1452--1461, 2015{\natexlab{b}}.

\bibitem[L{\"o}we et~al.(2022)L{\"o}we, Madras, Zemel, and
  Welling]{lowe2022amortized}
Sindy L{\"o}we, David Madras, Richard Zemel, and Max Welling.
\newblock Amortized causal discovery: Learning to infer causal graphs from
  time-series data.
\newblock In \emph{Conference on Causal Learning and Reasoning}, pp.\
  509--525. PMLR, 2022.

\bibitem[Mabrouk et~al.(2014)Mabrouk, Gonzales, Jabet-Chevalier, and
  Chojnacki]{mabrouk2014efficient}
Ahmed Mabrouk, Christophe Gonzales, Karine Jabet-Chevalier, and Eric Chojnacki.
\newblock An efficient {B}ayesian network structure learning algorithm in the
  presence of deterministic relations.
\newblock \emph{Frontiers in Artificial Intelligence and Applications},
  263:\penalty0 567--572, 2014.

\bibitem[Mitrovic et~al.(2018)Mitrovic, Sejdinovic, and
  Teh]{mitrovic2018causal}
Jovana Mitrovic, Dino Sejdinovic, and Yee~Whye Teh.
\newblock Causal inference via kernel deviance measures.
\newblock In \emph{Advances in Neural Information Processing Systems}, pp.\
  6986--6994, 2018.

\bibitem[Monti et~al.(2019)Monti, Zhang, and Hyvarinen]{monti2019causal}
Ricardo~Pio Monti, Kun Zhang, and Aapo Hyvarinen.
\newblock Causal discovery with general non-linear relationships using
  non-linear ica.
\newblock \emph{arXiv preprint arXiv:1904.09096}, 2019.

\bibitem[Mooij et~al.(2016)Mooij, Magliacane, and Claassen]{mooij2016joint}
Joris~M Mooij, Sara Magliacane, and Tom Claassen.
\newblock Joint causal inference from multiple contexts.
\newblock \emph{arXiv preprint arXiv:1611.10351}, 2016.

\bibitem[M{\"u}ller et~al.(2021)M{\"u}ller, Hollmann, Arango, Grabocka, and
  Hutter]{muller2021transformers}
Samuel M{\"u}ller, Noah Hollmann, Sebastian~Pineda Arango, Josif Grabocka, and
  Frank Hutter.
\newblock Transformers can do bayesian inference.
\newblock \emph{arXiv preprint arXiv:2112.10510}, 2021.

\bibitem[Pearl(2009)]{pearl2009causality}
Judea Pearl.
\newblock \emph{Causality}.
\newblock Cambridge university press, 2009.

\bibitem[Peters et~al.(2011)Peters, Mooij, Janzing, and
  Sch\"{o}lkopf]{Peters2011b}
J.~Peters, J.~M. Mooij, D.~Janzing, and B.~Sch\"{o}lkopf.
\newblock Identifiability of causal graphs using functional models.
\newblock In \emph{Proceedings of the 27th Annual Conference on {U}ncertainty
  in {A}rtificial {I}ntelligence ({UAI})}, pp.\  589--598, 2011.

\bibitem[Peters et~al.(2016)Peters, B{\"u}hlmann, and
  Meinshausen]{peters2016causal}
Jonas Peters, Peter B{\"u}hlmann, and Nicolai Meinshausen.
\newblock Causal inference by using invariant prediction: identification and
  confidence intervals.
\newblock \emph{Journal of the Royal Statistical Society: Series B (Statistical
  Methodology)}, 78\penalty0 (5):\penalty0 947--1012, 2016.

\bibitem[Rojas-Carulla et~al.(2018)Rojas-Carulla, Sch{\"o}lkopf, Turner, and
  Peters]{rojas2018invariant}
Mateo Rojas-Carulla, Bernhard Sch{\"o}lkopf, Richard Turner, and Jonas Peters.
\newblock Invariant models for causal transfer learning.
\newblock \emph{The Journal of Machine Learning Research}, 19\penalty0
  (1):\penalty0 1309--1342, 2018.

\bibitem[Sachs et~al.(2005)Sachs, Perez, Pe'er, Lauffenburger, and
  Nolan]{sachs2005causal}
Karen Sachs, Omar Perez, Dana Pe'er, Douglas~A Lauffenburger, and Garry~P
  Nolan.
\newblock Causal protein-signaling networks derived from multiparameter
  single-cell data.
\newblock \emph{Science}, 308\penalty0 (5721):\penalty0 523--529, 2005.

\bibitem[Scherrer et~al.(2021)Scherrer, Bilaniuk, Annadani, Goyal, Schwab,
  Sch{\"o}lkopf, Mozer, Bengio, Bauer, and Ke]{scherrer2021learning}
Nino Scherrer, Olexa Bilaniuk, Yashas Annadani, Anirudh Goyal, Patrick Schwab,
  Bernhard Sch{\"o}lkopf, Michael~C Mozer, Yoshua Bengio, Stefan Bauer, and
  Nan~Rosemary Ke.
\newblock Learning neural causal models with active interventions.
\newblock \emph{arXiv preprint arXiv:2109.02429}, 2021.

\bibitem[Shimizu et~al.(2006)Shimizu, Hoyer, Hyv{\"a}rinen, and
  Kerminen]{shimizu2006linear}
Shohei Shimizu, Patrik~O Hoyer, Aapo Hyv{\"a}rinen, and Antti Kerminen.
\newblock A linear non-gaussian acyclic model for causal discovery.
\newblock \emph{Journal of Machine Learning Research}, 7\penalty0
  (Oct):\penalty0 2003--2030, 2006.

\bibitem[Spiegelhalter \& Cowell(1992)Spiegelhalter and
  Cowell]{Spiegelhalter1992Learning}
D.~J. Spiegelhalter and R.~G. Cowell.
\newblock {Learning in probabilistic expert systems}.
\newblock \emph{Bayesian Statistics}, 4:\penalty0 447--466, 1992.

\bibitem[Spirtes et~al.(2000)Spirtes, Glymour, Scheines, Heckerman, Meek,
  Cooper, and Richardson]{spirtes2000causation}
Peter Spirtes, Clark~N Glymour, Richard Scheines, David Heckerman, Christopher
  Meek, Gregory Cooper, and Thomas Richardson.
\newblock \emph{Causation, prediction, and search}.
\newblock MIT press, 2000.

\bibitem[Squires et~al.(2020)Squires, Wang, and Uhler]{squires2020permutation}
Chandler Squires, Yuhao Wang, and Caroline Uhler.
\newblock Permutation-based causal structure learning with unknown intervention
  targets.
\newblock In \emph{Conference on Uncertainty in Artificial Intelligence}, pp.\
  1039--1048. PMLR, 2020.

\bibitem[Sun et~al.(2007)Sun, Janzing, Sch{\"o}lkopf, and
  Fukumizu]{sun2007kernel}
Xiaohai Sun, Dominik Janzing, Bernhard Sch{\"o}lkopf, and Kenji Fukumizu.
\newblock A kernel-based causal learning algorithm.
\newblock In \emph{Proceedings of the 24th international conference on Machine
  learning}, pp.\  855--862. ACM, 2007.

\bibitem[Tsamardinos et~al.(2006)Tsamardinos, Brown, and
  Aliferis]{tsamardinos2006max}
Ioannis Tsamardinos, Laura~E Brown, and Constantin~F Aliferis.
\newblock The max-min hill-climbing bayesian network structure learning
  algorithm.
\newblock \emph{Machine learning}, 65\penalty0 (1):\penalty0 31--78, 2006.

\bibitem[Vaswani et~al.(2017)Vaswani, Shazeer, Parmar, Uszkoreit, Jones, Gomez,
  Kaiser, and Polosukhin]{vaswani2017attention}
Ashish Vaswani, Noam Shazeer, Niki Parmar, Jakob Uszkoreit, Llion Jones,
  Aidan~N Gomez, {\L}ukasz Kaiser, and Illia Polosukhin.
\newblock Attention is all you need.
\newblock In \emph{Advances in neural information processing systems}, pp.\
  5998--6008, 2017.

\bibitem[Wang et~al.(2021{\natexlab{a}})Wang, King, Porcel, Kurth-Nelson, Zhu,
  Deck, Choy, Cassin, Reynolds, Song, et~al.]{wang2021alchemy}
Jane~X Wang, Michael King, Nicolas Porcel, Zeb Kurth-Nelson, Tina Zhu, Charlie
  Deck, Peter Choy, Mary Cassin, Malcolm Reynolds, Francis Song, et~al.
\newblock Alchemy: A structured task distribution for meta-reinforcement
  learning.
\newblock \emph{arXiv preprint arXiv:2102.02926}, 2021{\natexlab{a}}.

\bibitem[Wang et~al.(2021{\natexlab{b}})Wang, Du, Zhu, Ke, Chen, Hao, and
  Wang]{wang2021ordering}
Xiaoqiang Wang, Yali Du, Shengyu Zhu, Liangjun Ke, Zhitang Chen, Jianye Hao,
  and Jun Wang.
\newblock Ordering-based causal discovery with reinforcement learning.
\newblock \emph{arXiv preprint arXiv:2105.06631}, 2021{\natexlab{b}}.

\bibitem[Yoon et~al.(2018)Yoon, Jordon, and Van Der~Schaar]{yoon2018gain}
Jinsung Yoon, James Jordon, and Mihaela Van Der~Schaar.
\newblock Gain: Missing data imputation using generative adversarial nets.
\newblock \emph{arXiv preprint arXiv:1806.02920}, 2018.

\bibitem[Yu et~al.(2019)Yu, Chen, Gao, and Yu]{yu2019dag}
Yue Yu, Jie Chen, Tian Gao, and Mo~Yu.
\newblock Dag-gnn: Dag structure learning with graph neural networks.
\newblock \emph{arXiv preprint arXiv:1904.10098}, 2019.

\bibitem[Zhang et~al.(2012)Zhang, Peters, Janzing, and
  Sch{\"o}lkopf]{zhang2012kernel}
Kun Zhang, Jonas Peters, Dominik Janzing, and Bernhard Sch{\"o}lkopf.
\newblock Kernel-based conditional independence test and application in causal
  discovery.
\newblock \emph{arXiv preprint arXiv:1202.3775}, 2012.

\bibitem[Zheng et~al.(2018)Zheng, Aragam, Ravikumar, and Xing]{zheng2018dags}
Xun Zheng, Bryon Aragam, Pradeep~K Ravikumar, and Eric~P Xing.
\newblock {DAGs} with {NO TEARS}: Continuous optimization for structure
  learning.
\newblock In \emph{Advances in Neural Information Processing Systems}, pp.\
  9472--9483, 2018.

\bibitem[Zhu \& Chen(2019)Zhu and Chen]{zhu2019causal}
Shengyu Zhu and Zhitang Chen.
\newblock Causal discovery with reinforcement learning.
\newblock \emph{arXiv preprint arXiv:1906.04477}, 2019.

\end{thebibliography}
\bibliographystyle{iclr2023_conference}

\clearpage

\appendix
\section{Appendix}

\vspace{5mm}
\vskip3pt\hrule\vskip5pt
\localtableofcontents
\vskip3pt\hrule\vskip5pt
\clearpage

\subsection{Transformer Neural Networks}
\label{sec:app-transformers}

The transformer architecture, introduced in \citet{vaswani2017attention}, is a multi-layer neural network architecture using stacked self-attention and point-wise, fully connected, layers. 
The classic transformer architecture has an encoder and a decoder, but the encoder and decoder do not necessarily have to be used together. 

\paragraph{Scaled dot-product attention.}
The attention mechanism lies at the core of the transformer architecture. The transformer architecture uses a special form of attention, called the scaled dot-product attention. The attention mechanism allows the model to flexibility learn to weigh the inputs depending on the context. The input to the QKV attention consists of a set of queries, keys and value vectors. The queries and keys have the same dimensionality of $d_k$, and values often have a different dimensionality of $d_v$.  Transformers compute the dot products of the query with all keys, divide each by $\sqrt{d_k}$, and apply a softmax function to obtain the weights on the values. In practice, transformers compute the attention function on a set of queries simultaneously, packed together into a matrix $Q$. The keys and values are also packed together into matrices $K$ and $V$.  The matrix of outputs is computed as: $\mathrm{Attention}(Q, K, V) = \mathrm{softmax}(\frac{QK^T}{\sqrt{d_k}})V$.

\paragraph{Encoder.} The encoder is responsible for processing and summarizing the information in the inputs. The encoder is composed of a stack of $N$ identical layers, where each layer has two sub-layers. The first sub-layer consists of a multi-head self-attention mechanism, and the second is a simple, position-wise fully connected feed-forward network. Transformers employ a residual connection \citep{he2016deep} around each of the two sub-layers, followed by layer normalization \cite{ba2016layer}.  That is, the output of each sub-layer is $\mathrm{LayerNorm}(x + \mathrm{Sublayer}(x))$, where $\mathrm{Sublayer}(x)$ is the function implemented by the sub-layer itself.  

\paragraph{Decoder.} The decoder is responsible for transforming the information summarized by the encoder into the outputs. The decoder also composes of a stack of $N$ identical layers, with a small difference in the decoder transformer layer. In addition to the two sub-layers in each encoder layer, a decoder layer consists of a third sub-layer. The third sub-layer  performs  a multi-head attention over the output of the encoder stack.  Similar to the encoder, transformers employ residual connections around each of the sub-layers, followed by layer normalization.  Transformers also modify the self-attention sub-layer in the decoder stack to prevent positions from attending to subsequent positions.  This masking, combined with fact that the output embeddings are offset by one position, ensures that the predictions for position $i$ can depend only on the known outputs at positions less than $i$.

\subsection{Synthetic Data}
\label{appendix:synethtic_data}
In Section \ref{sec:CPDs}, we introduced the types of methods used to generate the conditional probability distribution. In this section, we discuss the details of these methods.

We evaluate our model on datasets generated from 5 different methods, which covers both continuous and discrete valued data generated with a varying degree of difficulty. For continuous data, we generated the data using three different methods: \textit{linear}, \textit{non-linear additive noise model} (ANM) and \textit{non-linear non-additive noise neural network model} (NN). For discrete data, we generated data using two different methods: \textit{MLP} and \textit{Dirichlet}. Let $X$ be a $N\times S$ matrix representing $S$ samples of a CBN with $N$ nodes and weighted adjacency matrix $A$, and let $Z$ be a random matrix of elements in $\mathcal{N}(0, 0.1)$. We describe each type of how different types of data is generated.

\begin{itemize}
    \item For \emph{linear data}, we follow the setup  in \cite{zheng2018dags} and \citet{yu2019dag}. Specifically, we generated data as $X_{n,:} = A_{n,:} X + Z_{n,:}$. The biases were initialized using $U[-0.5, 0.5]$, and the individual weights were initialized using a truncated normal distribution with standard deviation of $1.5$. For nodes with interventions, values are sampled from the uniform distribution $U[-1,1]$. 
    \item For \textit{additive noise models (ANM)} , we follow the setup in \citep{brouillard2020differentiable,lippe2021efficient}. We generated the data as $X_{n,:} = F_{n,:}(X) + 0.4 \cdot Z_{n,:}$, where $F$ is fully connected neural network, the weights are randomly initialized from $\mathcal{N}(0,1)$. The neural network has one hidden layer with $10$ hidden units, the activation function is a leaky relu with a slop of $0.25$. The noise variables are sampled from $\mathcal{N}(0, \sigma^2)$, where $\sigma^2\sim\mathcal{U}[1,2]$. For nodes with interventions, values are sampled from the uniform distribution $\mathcal{N}(2,1)$. 
    \item For \emph{non-additive noise neural network (NN) models} , we also follow the setup in \citep{brouillard2020differentiable,lippe2021efficient}. We generate the data as $X_{n,:} = F_{n,:}(X,Z_{n,:})$, such that $F$ is fully connected neural network, the weights are randomly initialized from $\mathcal{N}(0,1)$. The neural network has one hidden layer with $20$ hidden units, the activation function is a tanh function. The noise variables are sampled from $\mathcal{N}(0, \sigma^2)$, where $\sigma^2\sim\mathcal{U}[1,2]$. For nodes with interventions, values are sampled from the uniform distribution $\mathcal{N}(2,1)$.
    \item For \emph{MLP data}, the neural network has two fully connected layers of hidden dimensionality $32$. Following past work \citep{ke2020dependency,scherrer2021learning,lippe2021efficient}, we used a randomly initialized network. For nodes with interventions, values are randomly and independently sampled from ${U}\{1,\ldots,K\}$ where $K$ indicates the number of categories of the  discrete variable. 
    \item For Dirichlet data, the generator filled in the rows of a conditional probability table by sampling a categorical distribution from a Dirichlet prior with symmetric parameters $\alpha$. Values of $\alpha$ smaller than 1 encourage lower entropy distributions; values of $\alpha$ greater than 1 provide less information about the causal relationships among variables. Similar to MLP data, for nodes with interventions, values are randomly and independently sampled from ${U}\{1,\ldots,K\}$ where $K$ indicates the number of categories of the  discrete variable.
    
\end{itemize}

\subsection{Hyperparameters}
\label{appendix:hyperparam}

\begin{table}[h]
    \centering
    \caption{Hyperparameters used in all of our experiments.}
    \begin{tabular*}{.6\textwidth}{@{\extracolsep{\fill}}lc}
    \toprule
      \textbf{Hyperparameter} & \textbf{Value} \\ 
      \midrule
      % All activation fns & ReLU? \\ 
      Hidden state dimension & 128\\ 
      Encoder transformer layers & 10 \\ 
      Num. attention heads & 8 \\ 
      %Batch size  & 1 \\ 
      Optimizer & Adam\\ %\citep{kingma2014adam} \\
      Learning rate & \(1e^{-4}\) \\
      Number of random seeds & 3 \\
      $S$ (number of samples) & 1500 \\
      Training iterations & $50,000$ \\
      Num. training datasets $I$ & $15,000$ {\tiny($N\leq20$)}  
      $40,000$ {\tiny($N>20$)} \\
      \bottomrule
    \end{tabular*}
    \label{tab:app:hyper}
\end{table}

For all of our experiments (unless otherwise stated) our model was trained on $I=15,000$ (for graphs $N\leq20$) and on $I=40,000$ (for graphs $N>20$) pairs $\{(\calD^i,A^i)\}_{i=1}^{I}$, where each dataset $\calD^i$ contained $S=1500$ observational and interventional samples. For experiments on discrete data, a data-sample element $x^s$ could take values in $\{1, 2, 3 \}$. Details of the data generating process can be found in Section \ref{sec:CPDs}. For evaluation in Sections~\ref{sec:in_distribution} and  \ref{sec:ood_experiments}, our model was tested on $I'=128$ (different for the training) pairs $\{(\calD^{i'}, A^{i'})\}_{i'=1}^{I'}$, where each dataset $\calD^{i'}$ contained $S=1500$ observational and \textit{interventional} samples. For the Asia, Sachs and Child benchmarks, our model was still tested on $I'=128$ (different for the training) pairs $\{(\calD^{i'}, A^{i'})\}_{i'=1}^{I'}$, however, $A^{i'}=A^{j'}$ since there is only a single adjacency matrix in each one of the benchmarks.
%Each setting of the experiment was run with $3$ random seeds. 
%For each experimental setting, 
We present test results averaging performance over the $128$ datasets and $3$ random seeds and up to size $N=80$ graphs. The model was trained for $500,000$ iterations using the Adam optimizer \citep{kingma2014adam} with a learning rate of $1\mathrm{e}{-4}$. Also, refer to Table \ref{tab:app:hyper} for the list of hyperparameters presented in a table.

We parameterized our architecture such that inputs to the encoder were embedded into $128$-dimensional vectors. The encoder transformer has $10$ layers and $8$ attention-heads per layer. The final attention step for summarization has $8$ attention heads. The decoder was a smaller transformer with only $2$ layers and $8$ attention heads per layer.  Discrete inputs were encoded using an embedding layer before passing into our model.%\silvia{Already said? Or I am not undertanding.}

\subsection{Comparisons to Baselines}
\label{appendix:baseline_choices}

In Section \ref{sec:in_distribution}, we compare CSIvA to four strong baselines in the literature, ranging from classic causal discovery baselines to neural-network based causal discovery baselines. These baselines are DAG-GNN \citep{yu2019dag}, non-linear ICP \citep{heinze2018invariant},  DCDI \citep{brouillard2020differentiable} and ENCO \cite{lippe2021efficient}. 

Non-linear ICP, DCDI and ENCO can handle both observational and \textit{interventional} data, while DAG-GNN can only use observational data. Non-linear ICP could not scale to  graphs larger than 20, therefore we compare to DCDI \cite{brouillard2020differentiable} and ENCO \cite{lippe2021efficient} on larger graphs. All baselines are unsupervised methods, i.e. they are not tuned to a particular training dataset but instead rely on a general-purpose algorithm. 
We also compared to an all-absent model corresponding to a zero adjacency matrix, which acts as a sanity check baseline. 
We also considered other methods \citep{chickering2002optimal,hauser2012characterization,zhang2012kernel,gamella2020active}, but only presented a comparison with DCDI, ENCO, non-linear ICP and DAG-GNN as these have shown to be strong performing models in other works \citep{ke2020dependency,lippe2021efficient,scherrer2021learning}.

For Section \ref{sec:sim_to_real}, we also compared to additional baselines from \citet{chickering2002optimal,hauser2012characterization,zheng2018dags,gamella2020active,li2020supervised}. Note that methods from \citet{chickering2002optimal,zheng2018dags,yu2019dag,li2020supervised} take observational data only. 
DAG-GNN outputs several candidate graphs based on different scores, such as evidence lower bound or negative log likelihood, DCDI can also  be run in two different settings (DCDI-G and DCDI-DSF), we chose the best result to 
compare to our model. Note that non-linear ICP does not work on discrete data, i.e. on the MLP and Dirichlet data, therefore a small amount of Gaussian noise $\mathcal{N}(0,0.1)$ was added to this data in order for the method to run.

\subsection{Detailed results for Section \ref{sec:in_distribution}}
\label{appendix:in_distribution}
We present detailed results for experiments in Section \ref{sec:in_distribution} are described in the tables below. 

\subsubsection{Results on  linear data}
\label{appendix:linear_data}
Results for comparions between our model \CSIVA~ and baselines non-linear ICP \citep{heinze2018invariant}, DAG-GNN \citep{yu2019dag}, DCDI \citep{brouillard2020differentiable} and ENCO \citep{lippe2021efficient} are shown in Table \ref{table:linear_results} for smaller graphs with $N\leq 20$ and Table \ref{table:linear_larger_graphs_baselines} for larger graphs with $20<N\leq80$.

For smaller graphs with $N\leq 20$, all models that takes interventional data perform  significantly better compared to DAG-GNN \citep{yu2019dag}, which only takes observational data. 
\CSIVA~achieves Hamming distance ${\cal H}<7$ on  evaluated graphs up to size $20$. Similar to previous findings \citep{yu2019dag,ke2020dependency}, larger and denser graphs are more challenging to learn. Non-linear ICP achieves fairly good performance on smaller graphs ( $N\leq10$), however, the performance drops quickly as size of graphs increases ($N>10$). %at times approaching that of \CSIVA, 
Also, note that Non-linear ICP can not scale to graphs larger than $N>20$. It
also required a modification\footnote{Without this modification, the method achieved near chance performance.} to the dataset wherein multiple samples were collected from the same modified graph after a point intervention (20 samples per intervention), while other methods only sampled once per intervention.

For larger graphs of $20<N\leq280$, we compare to strongest baselines: DCDI and ENCO. Results are found in Table \ref{table:linear_larger_graphs_baselines}. \CSIVA~significantly outperforms both DCDI and ENCO. The difference because more apparent as the size of the graphs grow ($N\geq50$).

\begin{table*}[h]
\small
\centering  
\begin{tabular*}{\textwidth}{@{\extracolsep{\fill}}lcccccccc}
\toprule
 & \multicolumn{4}{c}{\textbf{ER = 1}} & \multicolumn{4}{c}{\textbf{ER = 2}} \\* \cmidrule(r){2-5} \cmidrule(l){6-9}
 
 & \multicolumn{1}{c}{\textbf{Var = 5}} & \multicolumn{1}{c}{\textbf{Var = 10}} & \multicolumn{1}{c}{\textbf{Var = 15}} & \multicolumn{1}{c}{\textbf{Var = 20}} 
 
 & \multicolumn{1}{c}{\textbf{Var = 5}} & \multicolumn{1}{c}{\textbf{Var = 10}} & \multicolumn{1}{c}{\textbf{Var = 15}} & \multicolumn{1}{c}{\textbf{Var = 20}}  \\
\midrule
Abs. & 2.50 &  5.00 & 7.50 & 10.00     & 5.00 &  10.00 & 15.00 & 20.00 \\
%{\bf \small \citet{zheng2018dags}} & ? & ? & ? & ? & ? & ? \\
{\small DAG-GNN} & 2.71 &  4.76& 7.71 & 11.32  & 5.20 & 8.81 &  17.81 & 22.21\\
{\small Non-linear ICP} & 0.47 & 1.10 & 6.3& 8.6         & 0.90 &  2.41 & 13.52 & 17.71 \\
{\small DCDI} & 0.23 & 0.78 & 4.51 & 6.21        & 0.92 &  2.01 & 8.12 & 12.59 \\
{\small ENCO} & 0.19 & 0.45 & 3.12 &   4.23      & 0.85 &  1.81 & 6.14 & 8.45 \\
Our Model & \textbf{0.12} \tiny{$\pm$ 0.03} & \textbf{0.35} \tiny{$\pm$ 0.05} & \textbf{2.01} \tiny{$\pm$ 0.07} & \textbf{2.21} \tiny{$\pm$ 0.07}& 
\textbf{0.81} \tiny{$\pm$ 0.05} & \textbf{1.73} \tiny{$\pm$ 0.04} &\textbf{5.32} \tiny{$\pm$ 0.19} &  \textbf{5.86} \tiny{$\pm$ 0.21}\\
%{\bf \small \citet{zheng2018dags}} & ? & ? & ? & ? & ? & ? \\
%{\bf  \small \cite{yu2019dag}} & 2.71 & 3.62 & 4.76 & 5.20 & 8.81 & 13.00 \\
%{\bf \small \cite{heinze2018invariant} } & 0.47 & 0.61 & 1.10 & 0.90 & 1.40 & 2.41 \\
%Our Model & \textbf{0.12} \tiny{$\pm$ 0.03} & \textbf{0.20} \tiny{$\pm$ 0.03} & \textbf{0.35} \tiny{$\pm$ 0.05} & \textbf{0.81} \tiny{$\pm$ 0.05} & \textbf{1.22} \tiny{$\pm$ 0.05}  & \textbf{1.73} \tiny{$\pm$ 0.04} \\
\bottomrule
\end{tabular*}
\caption{\textbf{Results on  Continuous linear data}. Hamming distance \calH~for learned and ground-truth edges on synthetic graphs, compared to other methods, averaged over 128 sampled graphs. The number of variables varies from 5 to 20, expected degree = 1 or 2, and the value of variables are drawn from $\mathcal{N}(0,0.1)$. Note that for \citep{heinze2018invariant}, the method required nodes to be causally ordered, and 20 repeated samples taken per intervention, as interventions were continuously valued. "Abs" baselines are All-Absent baselines, which is an baseline model that outputs all zero edges for the adjacency matrix. DCDI and ENCO are the strongest performing baselines.  \CSIVA~outperforms all baselines (including DCDI and ENCO).
% [This method also scales factorially with N, as it checks for all sets of possible parents, unless maxsetstocheck is set.]
}
\label{table:linear_results}
\end{table*}
%TODO: add the results for Gaussian Linear with std = 1.0
%We also conducted experiments on Continuous (Gaussian Linear data) with stand

\begin{table*}[h]
\centering  
{\small %
\begin{tabular}{l|c|c|c|c}
\toprule
{\bf Dataset } & {\bf All-absent} & {\bf DCDI} & {\bf ENCO} & {\bf CSIvA}  \\
\midrule
\small $N=30, ER = 1$ & 15.0 & 4.4 & 11.3 & {\bf 1.8} \\
\small $N=30, ER = 2$ & 30.0 & 9.2 & 29.1 & {\bf 5.6} \\
\small $N=50, ER = 1$ & 25.0 & 18.1 & 27.5 & {\bf 3.5} \\
\small $N=50, ER = 2$ & 50.0 & 23.7 & 73.2 & {\bf 11.7 } \\
\small $N=80, ER = 1$ & 40.0 & 29.5 & 39.2 & {\bf 9.8} \\
\small $N=80, ER = 2$ & 80.0 & 49.3 & 97.2 & {\bf 19.6} \\
% {\bf  \small ENCO \citep{}} & 2.2 & 3.2 & 5.3 & 4.7 & 6.1 & 9.3 \\
% {\bf \small{CSIvA}} & \textbf{0.26} \tiny{$\pm$ 0.05} & \textbf{0.83} \tiny{$\pm$ 0.06} & \textbf{2.37} \tiny{$\pm$ 0.07}  & \textbf{0.65} \tiny{$\pm$ 0.05}& \textbf{0.97} \tiny{$\pm$ 0.06} & \textbf{4.59} \tiny{$\pm$ 0.08} \\
\bottomrule
\end{tabular}
}
\caption{\textbf{Results on  Linear  data on larger graphs.} Hamming distance \calH~for learned and ground-truth edges on synthetic graphs, compared to other methods. The number of variables varies from 30 to 80, expected degree = 1 or 2. We  compare to DCDI \cite{brouillard2020differentiable} and ENCO \cite{lippe2021efficient} as they are the best performing baselines and they are able to handle larger graphs.
% [This method also scales factorially with N, as it checks for all sets of possible parents, unless maxsetstocheck is set.]
}
\label{table:linear_larger_graphs_baselines}
\end{table*}

\subsubsection{Results on  ANM data}
\label{appendix:anm_data}

For additive noise non-linear model (ANM) data, we compare to the strongest baseline models: DCDI and ENCO on $N\leq80$ graphs.  Results are found in Table \ref{table:anm_larger_graphs_baselines}. \CSIVA~achives hamming distance $\mathcal{H}<11$ on all graphs of size up to $80$. Therefore, \CSIVA~significantly outperforms strong baseline models DCDI and ENCO on all graphs. This difference becomes more apparent on larger graphs of $N\geq50$.

\begin{table*}[h]
\centering  
{\small %
\begin{tabular}{l|c|c|c|c}
\toprule
{\bf Dataset } & {\bf All-absent} & {\bf DCDI} & {\bf ENCO} & {\bf CSIvA}  \\
\midrule
\small $N=30, ER = 1$ & 15.0 &  3.1 & 14.7 & {\bf 0.7} \\
\small $N=30, ER = 2$ & 30.0 & 5.8  & 31.2 & {\bf 2.1} \\
\small $N=50, ER = 1$ & 25.0 & 7.8 & 25.6 & {\bf 0.7} \\
\small $N=50, ER = 2$ & 50.0 & 9.7 & 48.2 & {\bf 3.2 } \\
\small $N=80, ER = 1$ & 40.0 & 16.9 & 39.7 & {\bf 4.3} \\
\small $N=80, ER = 2$ & 80.0 & 37.1 & 78.2 & {\bf 10.6} \\
% {\bf  \small ENCO \citep{}} & 2.2 & 3.2 & 5.3 & 4.7 & 6.1 & 9.3 \\
% {\bf \small{CSIvA}} & \textbf{0.26} \tiny{$\pm$ 0.05} & \textbf{0.83} \tiny{$\pm$ 0.06} & \textbf{2.37} \tiny{$\pm$ 0.07}  & \textbf{0.65} \tiny{$\pm$ 0.05}& \textbf{0.97} \tiny{$\pm$ 0.06} & \textbf{4.59} \tiny{$\pm$ 0.08} \\
\bottomrule
\end{tabular}
}
\caption{\textbf{Results on continuous non-linear additive noise model (ANM) data with larger graphs.} Hamming distance \calH~for learned and ground-truth edges on synthetic graphs, compared to other methods. The number of variables varies from 30 to 80, expected degree = 1 or 2. We  compare to DCDI \cite{brouillard2020differentiable} and ENCO \cite{lippe2021efficient} as they are the best performing baselines and they are able to handle larger graphs.
% [This method also scales factorially with N, as it checks for all sets of possible parents, unless maxsetstocheck is set.]
}
\label{table:anm_larger_graphs_baselines}
\end{table*}

\subsubsection{Results on  NN data}
\label{appendix:nn_data}

Results for comparisons between \CSIVA~and strongest baselines DCDI and ENCO on non-additive noise non-linear (NN) data are found in Table \ref{table:nn_larger_graphs_baselines}. \CSIVA~achieves hamming distance $\mathcal{H}<11$ on all graphs. Hence, \CSIVA~significantly outperforms DCDI and ENCO on all graphs. The differences grow larger as the size of the graph grows.

\begin{table*}[h]
\centering  
{\small %
\begin{tabular}{l|c|c|c|c}
\toprule
{\bf Dataset } & {\bf All-absent} & {\bf DCDI} & {\bf ENCO} & {\bf CSIvA}  \\
\midrule
\small $N=30, ER = 1$ & 15.0 & 3.9 & 14.7 & {\bf 0.8} \\
\small $N=30, ER = 2$ & 30.0 & 6.7 &  29.2 & {\bf 2.1 } \\
\small $N=50, ER = 1$ & 25.0 & 9.4 &   23.5 & {\bf 1.3 } \\
\small $N=50, ER = 2$ & 50.0 & 10.4 & 49.5 & {\bf  3.3 } \\
\small $N=80, ER = 1$ & 40.0 & 19.5 &  39.2 & {\bf 5.3} \\
\small $N=80, ER = 2$ & 80.0 & 39.8 &  80.6& {\bf 10.5} \\
\bottomrule
\end{tabular}
}
\caption{\textbf{Results on continuous non-linear non-additive noise model (NN) data with larger graphs.} Hamming distance \calH~for learned and ground-truth edges on synthetic graphs, compared to other methods. The number of variables varies from 30 to 80, expected degree = 1 or 2. We  compare to DCDI \cite{brouillard2020differentiable} and ENCO \cite{lippe2021efficient} as they are the best performing baselines and they are able to handle larger graphs.
}
\label{table:nn_larger_graphs_baselines}
\end{table*}

\subsubsection{Results on MLP data}
Results for comparisons between our model \CSIVA~ and baselines non-linear ICP \citep{heinze2018invariant} and DAG-GNN \citep{yu2019dag} on MLP data are shown in Table \ref{table:synthetic_graphs}. MLP data is non-linear and hence more challenging compared to the continuous linear data. Our model \CSIVA~significantly outperforms non-linear ICP and DAG-GNN. The difference becomes more apparent as the graph size grows larger and more dense.

\begin{table*}[h]
\centering  
{\small %
\begin{tabular}{l|c|c|c|c}
\toprule
{\bf Dataset } & {\bf All-absent} & {\bf DCDI} & {\bf ENCO} & {\bf CSIvA}  \\
\midrule
\small $N=30, ER = 1$ & 15.0 & 11.1 & 12.5 & {\bf 8.0 } \\
\small $N=30, ER = 2$ & 30.0 & 17.4 & 16.9 & {\bf 15.6 } \\
\small $N=50, ER = 1$ & 25.0 & 27.3 & 28.3 & {\bf 17.6} \\
\small $N=50, ER = 2$ & 50.0 & 47.6 & 48.2 & {\bf  33.6 } \\
\small $N=80, ER = 1$ & 40.0 & 52.8  & 35.8 & {\bf 28.6} \\
\small $N=80, ER = 2$ & 80.0 & 75.3 &  61.5 & {\bf 52.3} \\
\bottomrule
\end{tabular}
}
\caption{\textbf{Results on discrete MLP data with larger graphs.} Hamming distance \calH~for learned and ground-truth edges on synthetic graphs, compared to other methods. The number of variables varies from 30 to 80, expected degree = 1 or 2. We  compare to DCDI \cite{brouillard2020differentiable} and ENCO \cite{lippe2021efficient} as they are the best performing baselines and they are able to handle larger graphs.
}
\label{table:mlp_larger_graphs_baselines}
\end{table*}

\begin{figure}
     \centering
    \begin{subfigure}[t]{0.45\textwidth}
        \raisebox{-\height}{\includegraphics[width=\textwidth]{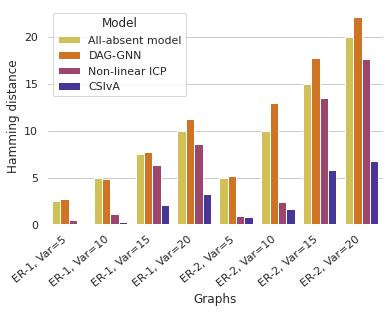}}
        \caption{\textbf{Results on Linear data on $N\leq20$ graphs.}}
    \end{subfigure}
    % \hfill
    \begin{subfigure}[t]{0.45\textwidth}
        \raisebox{-\height}{\includegraphics[width=\textwidth]{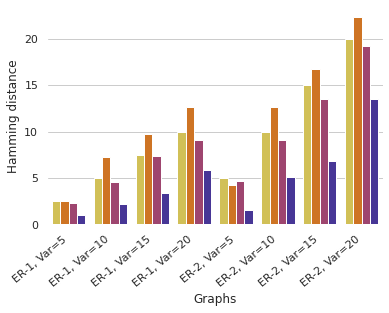}}
        \caption{\textbf{Results on MLP data  on $N\leq20$ graphs.}}
    \end{subfigure}
    %%%%%%%%%%%%%%%%%%%%%%%%%%%%%%%%%%%%second row
 \caption{Hamming distance \calH~between predicted and ground-truth adjacency matrices on the Linear and MLP data  for $N\leq20$ graphs, compared to DAG-GNN \citep{yu2019dag} and non-linear ICP \citep{heinze2018invariant}, averaged over 128 sampled graphs. 
 \CSIVA~significantly outperforms all other baselines.}  
 \label{fig:continous_mlp_results}
\end{figure}

\begin{table*}[h]
\small
\centering  
\begin{tabular*}{\textwidth}{@{\extracolsep{\fill}}lcccccccc}
\toprule
 & \multicolumn{4}{c}{\textbf{ER = 1}} & \multicolumn{4}{c}{\textbf{ER = 2}} \\* \cmidrule(r){2-5} \cmidrule(l){6-9}
 
 & \multicolumn{1}{c}{\textbf{Var = 5}} & \multicolumn{1}{c}{\textbf{Var = 10}} & \multicolumn{1}{c}{\textbf{Var = 15}} & \multicolumn{1}{c}{\textbf{Var = 20}}
 
 & \multicolumn{1}{c}{\textbf{Var = 5}} & \multicolumn{1}{c}{\textbf{Var = 10}} & \multicolumn{1}{c}{\textbf{Var = 15}} & \multicolumn{1}{c}{\textbf{Var = 20}} \\
\midrule
Abs. & 2.50 & 5.00 & 7.50 & 10.00 & 5.00 & 10.00  & 15.00 & 20.00 \\
{\bf  \small DAG-GNN} & 2.52 & 7.30 & 9.74  &  12.72      & 4.33 & 12.78 & 16.73  & 22.33\\
{\bf \small Non-linear ICP } & 2.43 & 4.62 & 7.42    &   9.05     & 4.76 & 9.12 & 13.52 & 19.25 \\
CSIvA & \textbf{0.98} \tiny{$\pm$ 0.16} & \textbf{2.25} \tiny{$\pm$ 0.17}&     \textbf{3.38} \tiny{$\pm$ 0.12} &\textbf{5.92} \tiny{$\pm$ 0.19}                                     & \textbf{1.51}\tiny{$\pm$ 0.47} &  \textbf{5.12} \tiny{$\pm$ 0.26} & \textbf{6.82} \tiny{$\pm$ 0.23} & \textbf{13.50} \tiny{$\pm$ 0.35}  \\
\bottomrule
\end{tabular*}
\caption{\textbf{Results on  MLP data}. Hamming distance \calH~for learned and ground-truth edges on synthetic graphs, compared to other methods, averaged over 128 sampled graphs ($\pm$ standard deviation). The number of variables varies from 5 to 20, expected degree = 1 or 2, and the dimensionality of the variables are fixed to 3. 
We compared to DAG-GNN  \citep{yu2019dag}, which is a  strong baseline that uses observational data. We also compare to Non-linear ICP  \citep{heinze2018invariant}, which is a strong baseline that uses \textit{interventional} data. Note that for \citep{heinze2018invariant}, the method required nodes to be causally ordered, and Gaussian noise $\mathcal{N}(0, 0.1)$ to be added. "Abs" baselines are All-Absent baselines, which is an baseline model that outputs all zero edges for the adjacency matrix.
}
\vspace{-1\baselineskip}
\label{table:synthetic_graphs}
\end{table*}

%%%%%%%%%%%%%%%%%%%%%%%%%%%%%%%%%%%%%%%%%%%%%%%%%%%%%%%%%%%%%%%%%%%%%%%%%%%%%%%
%%%%%%%%%%%%%%%%%%%%%%%%%%%%%%%%%%%%%%%%%%%%%%%%%%%%%%%%%%%%%%%%%%%%%%%%%%%%%%%
\subsubsection{Results on Dirichlet data.}

We run two sets of experiments on Dirichlet data. The first set of experiments is aimed at understanding how different $\alpha$ values impact the performance of our model. In the second set of experiments, we compare the performance of our model to four strong baselines: DCDI \cite{brouillard2020differentiable}, ENCO \cite{lippe2021efficient}, non-linear ICP \cite{heinze2018invariant} and DAG-GNN \cite{yu2019dag}.

The results for the first set of experiments are shown in Table \ref{table:dirichlet_data_all_alpha}.  Our model performs well on all graphs where $\alpha\leq 0.5$, and the performance starts to degrading as $\alpha=1.0$. When $\alpha=5.0$, our model is almost performing similarly to the All-absent model (outputting all zero edges). This is to be expected, as larger alpha values is less informative of the causal relationships between variables.

\begin{wrapfigure}{r}{0.45\textwidth}
   % \vspace{-1.5\baselineskip}
    \includegraphics[width=0.42\textwidth]{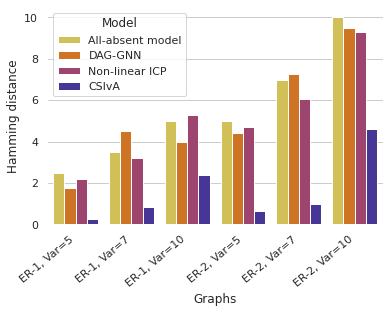}
    \caption{\textbf{Results on Dirichlet data on $N\leq20$ graphs}. Hamming distance \calH~~between predicted and ground-truth adjacency matrices on  Dirichlet data, averaged over 128 sampled graphs. 
    }
    \label{fig:dirichlet_results}
\end{wrapfigure}

\begin{table*}[h]
\centering  
\begin{tabular*}{\textwidth}{@{\extracolsep{\fill}}lcccccc}
\toprule
 & \multicolumn{3}{c}{\textbf{ER = 1}} & \multicolumn{3}{c}{\textbf{ER = 2}} \\* \cmidrule(r){2-4} \cmidrule(l){5-7}
 & \multicolumn{1}{c}{\textbf{Var = 5}} & \multicolumn{1}{c}{\textbf{Var = 7}} & \multicolumn{1}{c}{\textbf{Var = 10}} & \multicolumn{1}{c}{\textbf{Var = 5}} & \multicolumn{1}{c}{\textbf{Var = 7}} & \multicolumn{1}{c}{\textbf{Var = 10}} \\
\midrule
All-absent baseline & 2.5 & 3.5 & 5.0 & 5.0 & 7.0 & 10.0 \\
{\bf \small \citep{yu2019dag}} & 1.75 & 4.5 & 4.0 & 4.5 & 7.25 & 9.50 \\
{\bf  \small \citep{heinze2018invariant}} & 2.2 & 3.2 & 5.3 & 4.7 & 6.1 & 9.3 \\
{\bf \small{CSIvA}} & \textbf{0.26} \tiny{$\pm$ 0.05} & \textbf{0.83} \tiny{$\pm$ 0.06} & \textbf{2.37} \tiny{$\pm$ 0.07}  & \textbf{0.65} \tiny{$\pm$ 0.05}& \textbf{0.97} \tiny{$\pm$ 0.06} & \textbf{4.59} \tiny{$\pm$ 0.08} \\
\bottomrule
\end{tabular*}
\caption{\textbf{Results on Dirichlet data.} Hamming distance \calH~for learned and ground-truth edges on synthetic graphs, compared to other methods, averaged over 128 sampled graphs ($\pm$ standard deviation). The number of variables varies from 5 to 10, expected degree = 1 or 2,  the dimensionality of the variables are fixed to 3, and the $\alpha$ is fixed to $1.0$. We compare to the strongest causal-induction methods that uses observational data \citep{yu2019dag} and the strongest that uses \textit{interventional} data \citep{heinze2018invariant}.
}
\label{table:dirichlet_data_baselines}
\end{table*}

For the second set of experiments, we compared \CSIVA~to non-linear ICP and DAG-GNN on graphs of size up to 20. To limit the number of experiments to run, we set $\alpha=1.0$, as this gives the least amount of prior information to \CSIVA. As shown in Figure \ref{fig:dirichlet_results}, our model significantly outperforms non-linear ICP and DAG-GNN. Our model achieves ${\mathcal H}<5$ on size $10$ graphs, almost half of the error rate compared to non-linear ICP and DAG-GNN,  both achieving a significantly higher Hamming distance (${\mathcal H}=9.3$ and ${\mathcal H}=9.5$ respectively) on larger and denser graphs. Refer to Table \ref{table:dirichlet_data_baselines} for complete sets of results.

For larger graphs ($N>20$), we compare \CSIVA~to DCDI \cite{brouillard2020differentiable}
 and ENCO \cite{lippe2021efficient} as they could scale to larger graphs and are the strongest baselines. The results are illustrated in Figure \ref{fig:dirichlet_results_large} and the detailed results are found in Table \ref{table:dirichlet_larger_graphs_baselines}. As the graphs get larger, the performance of baselines DCDI and ENCO drops significantly, for dense and largr graphs (ER-2, $N=80$), the baseline models achieve almost near all-absent performance, while our model performs significantly better (achieving almost $30\%$ performance gain in terms of structured hamming distance).
 
\begin{table*}[h]
\small
\centering  
\begin{tabular*}{\textwidth}{@{\extracolsep{\fill}}lcccccccc}

\toprule
 & \multicolumn{4}{c}{\textbf{ER = 1}} & \multicolumn{4}{c}{\textbf{ER = 2}} \\* \cmidrule(r){2-5} \cmidrule(l){6-9}
 & \multicolumn{1}{c}{\textbf{Var = 5}} & %\multicolumn{1}{c}{\textbf{Var = 7}} &
 \multicolumn{1}{c}{\textbf{Var = 10}} & \multicolumn{1}{c}{\textbf{Var = 15}} &
 \multicolumn{1}{c}{\textbf{Var = 20}} &\multicolumn{1}{c}{\textbf{Var = 5}} & %\multicolumn{1}{c}{\textbf{Var = 7}} &
 \multicolumn{1}{c}{\textbf{Var = 10}} & \multicolumn{1}{c}{\textbf{Var = 15}} 
 & \multicolumn{1}{c}{\textbf{Var = 20}} \\
\midrule
{\textbf\small $\alpha=0.1$} & 0.18\tiny{$\pm$ 0.03} & 0.72\tiny{$\pm$ 0.04} & 1.31\tiny{$\pm$ 0.04} & 2.45 \tiny{$\pm$ 0.04}& 0.39\tiny{$\pm$ 0.04} &  1.27\tiny{$\pm$ 0.07} & 1.98\tiny{$\pm$ 0.12} & 4.09\tiny{$\pm$ 0.04}\\

{\textbf\small $\alpha=0.25$} & 0.14\tiny{$\pm$ 0.03} & 0.77\tiny{$\pm$ 0.05} & 1.62\tiny{$\pm$ 0.05} & 3.51 \tiny{$\pm$ 0.05}& 0.29\tiny{$\pm$ 0.04}  & 1.27\tiny{$\pm$ 0.07}  & 3.04\tiny{$\pm$ 0.20} & 6.41\tiny{$\pm$ 0.12}\\
{\bf  \small  $\alpha=0.5$} & 0.14\tiny{$\pm$ 0.04} & 0.94\tiny{$\pm$ 0.05} & 4.26\tiny{$\pm$ 0.07} & 7.35\tiny{$\pm$ 0.04} & 0.41 \tiny{$\pm$ 0.03} & 2.11\tiny{$\pm$ 0.06} & 8.25\tiny{$\pm$ 0.07}  & 15.54\tiny{$\pm$ 0.10} \\
{\bf \small $\alpha= 1.0$} & 0.26\tiny{$\pm$ 0.05}  & 2.37\tiny{$\pm$ 0.07} & 4.90\tiny{$\pm$ 0.05} & 10.10\tiny{$\pm$ 0.07} & 0.68\tiny{$\pm$ 0.03} &  4.32\tiny{$\pm$ 0.07} & 10.24\tiny{$\pm$ 0.07} & 21.81\tiny{$\pm$ 0.07}\\
{\bf  \small  $\alpha= 5.0$} & 1.27\tiny{$\pm$ 0.12}  & 4.9\tiny{$\pm$ 0.05} & 14.73\tiny{$\pm$ 0.11} & 19.49\tiny{$\pm$ 0.05} & 3.21\tiny{$\pm$ 0.05}  & 9.99 \tiny{$\pm$ 0.03}& 24.19\tiny{$\pm$ 0.05} & 37.03\tiny{$\pm$ 0.24} \\
Abs. & 2.5 & 5.0 & 7.5 & 10.0 & 5.0  & 10.0 & 15.0 & 20.0\\
%{\bf \small $\alpha=0.1$} & 0.18 & 0.37 & 0.72 & 0.39 &  0.84 & 1.27 \\
%{\bf \small $\alpha=0.25$} & 0.14 & 0.41 & 0.77 & 0.29 & 0.64 & 1.27 \\
%{\bf  \small  $\alpha=0.5$} & 0.14 & 0.43 & 0.94 & 0.41 & 0.79 & 2.11 \\
%{\bf \small $\alpha= 1.0$} & 0.27 & 0.63 & 2.31  & 0.68 & 1.22 & 4.32 \\
%{\bf  \small  $\alpha= 5.0$} & 1.27 & 2.56 & 4.91 &  3.21 & 7.0  & 9.99 \\
%All-absent Model & 2.5 & 3.5 & 5.0 & 5.0 & 7.0 & 10.0 \\
\bottomrule
\end{tabular*}
\caption{\textbf{Results on Dirichlet data.} Hamming distance \calH~ (lower is better) for learned and ground-truth edges on synthetic graphs, averaged over 128 sampled graphs. Our model accomplished a hamming distance of less than $2.5$ for Dirichlet data with $\alpha <= 0.5$. "Abs" baselines are All-Absent baselines, which is an baseline model that outputs all zero edges for the adjacency matrix.
% The number of variables varies from 5 to 10, expected degree = 1 or 2, and the dimensionality of the variables are fixed to 3.
% [This method also scales factorially with N, as it checks for all sets of possible parents, unless maxsetstocheck is set.]
}
\label{table:dirichlet_data_all_alpha}
\end{table*}

\begin{table*}[h]
\centering  
{\small %
\begin{tabular}{l|c|c|c|c}
\toprule
{\bf Dataset } & {\bf All-absent} & {\bf DCDI} \cite{brouillard2020differentiable} & {\bf ENCO} \cite{lippe2021efficient} & {\bf CSIvA}  \\
\midrule
\small $N=30, ER = 1$ & 15.0 & 8.3 & 4.9 & {\bf 2.9} \\
\small $N=30, ER = 2$ & 30.0 & 12.3 & 8.1 & {\bf 6.5} \\
\small $N=40, ER = 1$ & 20.0 & 16.2 & 9.9 & {\bf 5.3} \\
\small $N=40, ER = 2$ & 40.0 & 21.6 & 14.3 & {\bf 9.6} \\
\small $N=50, ER = 1$ & 25.0 & 26.4 & 17.7 & {\bf 8.9} \\
\small $N=50, ER = 2$ & 50.0 & 47.6 & 28.6 & {\bf 17.3} \\
\small $N=60, ER = 1$ & 30.0 & 28.4 & 25.6   & {\bf 12.6} \\
\small $N=60, ER = 2$ & 60.0 & 57.6   & 38.2 & {\bf 26.6} \\
\small $N=70, ER = 1$ & 35.0 & 34.1 & 29.1 & {\bf 18.4} \\
\small $N=70, ER = 2$ & 70.0 & 71.2 & 53.4 & {\bf 40.9} \\
\small $N=80, ER = 1$ & 40.0 & 54.1 & 39.1 & {\bf 23.1} \\
\small $N=80, ER = 2$ & 80.0 & 78.6 & 72.6 & {\bf 56.3} \\
% {\bf  \small ENCO \citep{}} & 2.2 & 3.2 & 5.3 & 4.7 & 6.1 & 9.3 \\
% {\bf \small{CSIvA}} & \textbf{0.26} \tiny{$\pm$ 0.05} & \textbf{0.83} \tiny{$\pm$ 0.06} & \textbf{2.37} \tiny{$\pm$ 0.07}  & \textbf{0.65} \tiny{$\pm$ 0.05}& \textbf{0.97} \tiny{$\pm$ 0.06} & \textbf{4.59} \tiny{$\pm$ 0.08} \\
\bottomrule
\end{tabular}
}
\caption{\textbf{Results on Dirichlet data with larger graphs.} Hamming distance \calH~for learned and ground-truth edges on synthetic graphs, compared to other methods. The number of variables varies from 30 to 80, expected degree = 1 or 2, and the $\alpha$ is fixed to $0.1$. We  compare to the approaches from \cite{brouillard2020differentiable} and \cite{lippe2021efficient} as they are the strongest baselines and they are able to handle larger graphs.
% [This method also scales factorially with N, as it checks for all sets of possible parents, unless maxsetstocheck is set.]
}
\label{table:dirichlet_larger_graphs_baselines}
\end{table*}

\begin{table*}[h]
\centering  
\begin{tabular*}{\textwidth}{@{\extracolsep{\fill}}lcccccc}
\toprule
 & \multicolumn{3}{c}{\textbf{ER = 1}} & \multicolumn{3}{c}{\textbf{ER = 2}} \\* \cmidrule(r){2-4} \cmidrule(l){5-7}
 & \multicolumn{1}{c}{\textbf{Var = 10}} & \multicolumn{1}{c}{\textbf{Var = 15}} & \multicolumn{1}{c}{\textbf{Var = 20}} & \multicolumn{1}{c}{\textbf{Var = 10}} & \multicolumn{1}{c}{\textbf{Var = 15}} & \multicolumn{1}{c}{\textbf{Var = 20}} \\
\midrule
All-absent baseline & 5.0 & 7.5 & 10.0 & 10.0 & 15.0 & 20.0 \\
\midrule
{\textbf\small $\alpha=0.1$} & 0.5 & 1.2  & 2.4 & 0.9 &  2.3 & 3.4\\
{\textbf\small $\alpha=0.25$} & 0.5 & 1.7 & 3.3   & 1.7&  3.5 & 4.6 \\
\bottomrule
\end{tabular*}
\caption{\textbf{Results on imperfect interventions.} Hamming distance \calH~for learned and ground-truth edges on synthetic graphs, compared to other methods, averaged over 128 sampled graphs ($\pm$ standard deviation). The number of variables varies from 10 to 20, expected degree = 1 or 2,  the dimensionality of the variables are fixed to 3, and the $\alpha$ is fixed to $0.1$ or $0.25$.
}
\label{table:imperfect_interventions}
\end{table*}

\subsubsection{Soft interventions}
\label{sec:imperfect_interventions}
In addition to hard interventions, we also consider soft interventions. Hence, we further evaluate our model on  imperfect interventions as discussed in Section \ref{sec:in_distribution}. To limit the number of experiments to run, we focus on Dirichlet data with $\alpha=0.1$ and $\alpha=0.25$. 

Soft interventions on Dirichlet data is generated as follows. For an intervention  on variable $X_i$, we first sample the $\alpha_i$ for the Dirichlet distribution from a Uniform distribution ${U}\{0.1,0.3, 0.5, 0.7,0.9\}$, then using the sampled $\alpha_i$, we sample the conditional probabilities from the Dirichlet distribution with the new $\alpha_i$. 

The results for \CSIVA~on soft interventions are shown in Table \ref{table:imperfect_interventions}. \CSIVA~ was able to achieve a hamming distance $\mathcal{H}<5$ on all graphs of size $H\leq20$. These results are strong indications that that our model still works well on the imperfect interventions.

\begin{table*}[h]
\centering  
{\small
\begin{tabular*}{\textwidth}{@{\extracolsep{\fill}}lcccccccc}
\toprule
 & \multicolumn{4}{c}{\textbf{ER = 1}} & \multicolumn{4}{c}{\textbf{ER = 2}} \\* \cmidrule(r){2-5} \cmidrule(l){6-9}
 & \multicolumn{2}{c}{\textbf{Var = 15}} & \multicolumn{2}{c}{\textbf{Var = 20}} & \multicolumn{2}{c}{\textbf{Var = 15}} & \multicolumn{2}{c}{\textbf{Var = 20}} \\* \cmidrule(r){2-3} \cmidrule(l){4-5} \cmidrule(l){6-7} \cmidrule(l){8-9} 
 & \multicolumn{1}{c}{\textbf{\small $\alpha=0.1$}} & 
 \multicolumn{1}{c}{\textbf{\small $\alpha=0.25$}} &
 \multicolumn{1}{c}{\textbf{\small $\alpha=0.1$}} & 
 \multicolumn{1}{c}{\textbf{\small $\alpha=0.25$}} & 
 \multicolumn{1}{c}{\textbf{\small $\alpha=0.1$}} & 
 \multicolumn{1}{c}{\textbf{\small $\alpha=0.25$}} & 
 \multicolumn{1}{c}{\textbf{\small $\alpha=0.1$}} & 
 \multicolumn{1}{c}{\textbf{\small $\alpha=0.25$}} \\
\midrule
{\small Known} & 1.3 & 1.7 & 2.5 & 3.4 & 2.3 & 3.0 & 4.1 & 6.3 \\
{\small Unknown} & 1.4 & 1.9 & 2.5 & 3.3 & 2.3 & 3.5 & 4.1 & 6.6 \\
\bottomrule
\end{tabular*}
}
\caption{\textbf{Results on unknown interventions compared to known interventions.} Hamming distance \calH~from learned and ground-truth edges on synthetic graphs, for known vs unknown interventions, averaged over 128 sampled graphs ($\pm$ standard deviation). The number of variables varies from 15 to 20, expected degree = 1 or 2, $\alpha$ = $0.1$ or $0.25$, and the dimensionality of the variables are fixed to 3.}
\label{table:unknown_interventions}
\end{table*}

\subsubsection{Unknown Interventions}
\label{sec:appendix_unknown_intervention}
Until now, we have considered cases where the target of the intervention is known, we now consider the case when the target of the intervention is unknown, this is also referred to as unknown interventions. Again, to limit the number of experiments to run, we focus on Dirichlet data with $\alpha=0.1$ and $\alpha=0.25$. The model does not know the target of the intervention and all other training procedures remains exactly the same. The results are shown in Table \ref{table:unknown_interventions}. We compare how well \CSIVA performs on known and unknown interventions. We can see that the performance for the known and unknown interventions are almost the same for sparser graphs (ER-1). The differences increases slightly for denser graphs with higher values of $\alpha$, for example $ER-2$ graphs with $\alpha=0.25$. This is to be expected, as denser graphs with higher $\alpha$ values are more challenging to learn. The biggest differences for the performances of known and unknown interventions is less than $\mathcal{H}<1.0$ in hamming distance. These results shown clear indications that our model performs well even when the intervention target is unknown.

\subsection{Detailed results for Out-of-distribution experiments}
\label{sec:appendix_ood_exp}

This section contains detailed results for the out-of-distribution experiments experiments in Section \ref{sec:ood_experiments}. Results for the experiments on varying graph density in Section \ref{sec:ood_experiments} are shown in Table \ref{table:ood_generalization_mlp} and  Table \ref{table:ood_generalization_ER_dirichlet}. The results on varying $\alpha$ values in Section \ref{sec:ood_experiments} are found in Table \ref{table:ood_generalization_alpha}. Our model generalizes well to test OOD test distributions, where either the graphs can vary in terms of sparsity or the conditional probability can vary in terms of $\alpha$ values for the Dirichlet distributions.

\begin{table}
\parbox{.45\linewidth}{
\centering
\begin{tabular}{llrrr}
\toprule
{\bf \small Train} & & {\small ER-1}& {\small ER-2} & {\small ER-3} \\ 
\midrule
&                   {\small ER-1}   &  {1.2} &  \colorbox{lightmintbg}{0.9}  & 1.3  \\
{\bf \small Test} & {\small ER-2}   &  3.3 &  \colorbox{lightmintbg}{1.8} &  2.1   \\
&                   {\small ER-3}   &  5.0 &  \colorbox{lightmintbg}{2.8}  & {2.8}    \\
\bottomrule
\end{tabular}
\vspace{1\baselineskip}
\caption{\textbf{Results on varying graph density for MLP data:} Hamming distance \calH~ between predicted and ground-truth adjacency matrices.}
\label{table:ood_generalization_mlp}
}
\hfill
\parbox{.45\linewidth}{
\centering
\begin{tabular}{llrrr}
\toprule
{\bf \small Train} & & {\small ER-1}& {\small ER-2} & {\small ER-3} \\ 
\midrule
                  &  {\small ER-1} &  \colorbox{lightmintbg}{0.19} & 0.21     & 0.28 \\
{\bf \small Test} & {\small }  & 0.86  &  0.29 &  \colorbox{lightmintbg}{0.25}  \\
                  & {\small ER-3} &   1.61 &  0.60 & \colorbox{lightmintbg}{0.23}    \\
\bottomrule
\end{tabular}
\vspace{1\baselineskip}
\caption{\textbf{Results on graph sparsity for Dirichlet data ($\alpha=1$):} Hamming distance \calH~~between predicted and ground-truth adjacency matrices.}
\label{table:ood_generalization_ER_dirichlet}
}
\end{table}

\begin{table}
% \vspace{-1\baselineskip}
\centering
\begin{tabular}{llrrr}
\toprule
{\bf \small Train} & & {\small $\alpha=0.1$} &{\small $\alpha=0.25$}   &{\small $\alpha=0.5$}  \\ 
\midrule
&                   {\small $\alpha=0.1$ }   &  \colorbox{lightmintbg}{0.31} &  0.33  & 0.52  \\
{\bf \small Test} & {\small $\alpha=0.25$ }   &  0.72 &  \colorbox{lightmintbg}{0.40} &  0.41   \\
&                   {\small $\alpha=0.5$ }   &  1.8 &  0.71  & \colorbox{lightmintbg}{0.35}    \\
\bottomrule
\end{tabular}
\vspace{1\baselineskip}
\caption{\textbf{Results on varying $\alpha$ values for Dirichlet data:} Hamming distance \calH~between predicted and ground-truth adjacency matrices.}
\label{table:ood_generalization_alpha}
\end{table}

\subsection{Ablation studies and analyses}
\label{sec:appendix_ablations}
In order to better understand the performance of our model, we performed further analysis and ablations. This section contains full results for  Section \ref{sec:analysis}. In this section, we aim to answer the following questions: 
\begin{itemize}
    \item What does visualizations of the generated graphs look like?
    \item Are these generated graphs from our model acyclic?
    \item Does intervention data improve identifiability in our model?
    \item How does varying the number of samples impact the performance of our model?
    \item How does different components of our model (such as sample-level attention and auxiliary loss) impact the performance of our model?
\end{itemize}

\subsubsection{Visualization of sampled graphs}

\begin{wrapfigure}{r}{0.5\textwidth}
    \begin{center}
    \vspace{-1\baselineskip}
    \includegraphics[scale=0.5]{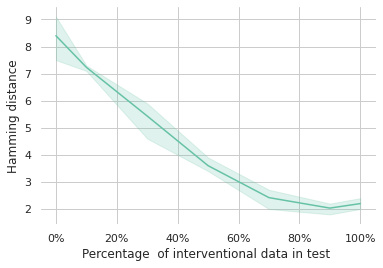}
    \end{center}
    \caption{\small \textbf{Results on varying amount of interventions in data.} Hamming distance \calH~for learned and ground-truth edges on synthetic Dirichlet graphs ($Var=15$ and $\alpha=0.25$). Pure observational data ($0\%$ interventions) performs the worst, which is to be expect, since intervention data is need for causal identifiability. The performance of our model improves as it observes more interventions, this suggest that our model is able to extract useful information from interventions in order to predict the causal structure.}
    \vspace{-1\baselineskip}
    \label{fig:percent_intervention}
\end{wrapfigure}
\begin{figure}
\centering
%\begin{subfigure}[b]{1\linewidth}
%{
\includegraphics[width=0.80\linewidth]{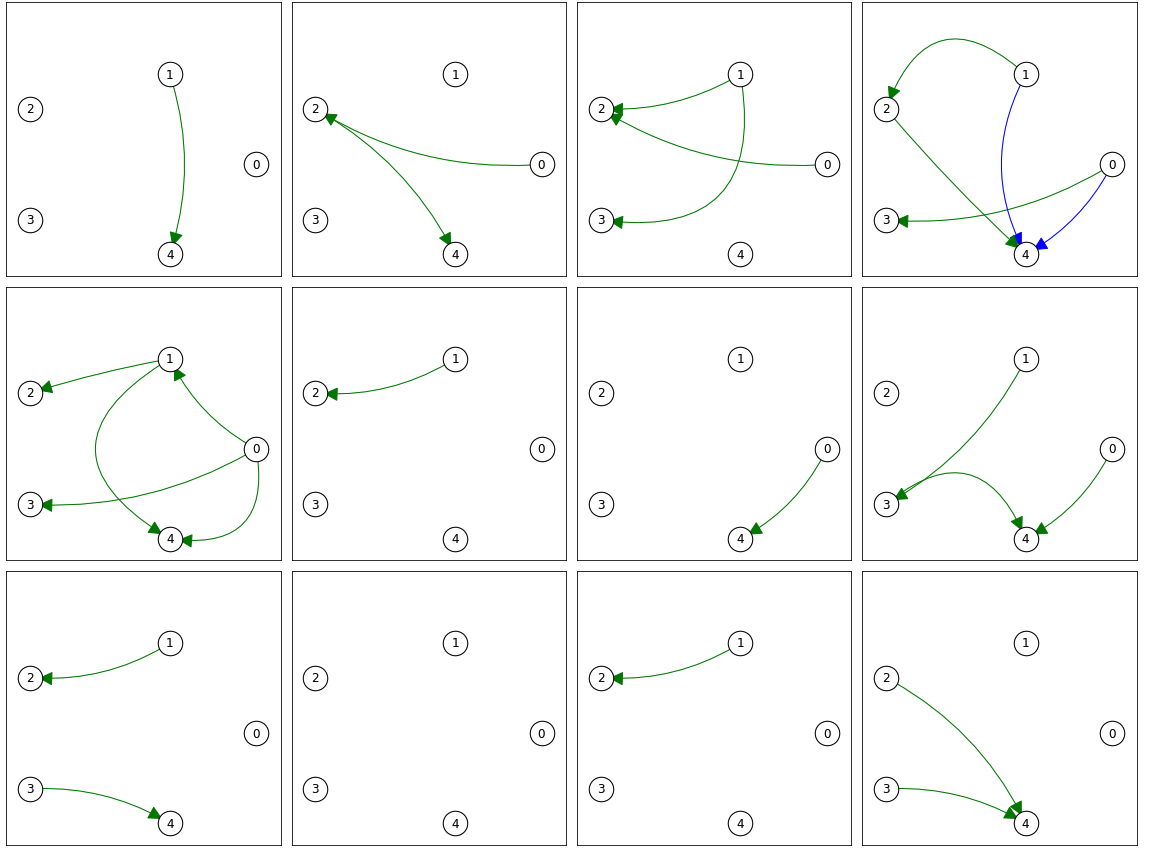}
\caption{This figures visualizes samples that our model generated on test data. The model was trained and tested on MLP data of size $5$ with  \textbf{ER-$1$} graphs. The samples are randomly chosen. The green edges indicate that our model has generated the correct edges; red edges indicate edges that our model had missed; and blue edges are the ones that our model  generated, which were not in the groundtruth graph. As shown above, our model is able to generate the correct graph almost all of the times, while only occasionally generating 1 or 2 incorrect edges in a graph.}
\label{fig:samples_er1}
\end{figure}
\begin{figure}
\centering
\includegraphics[width=0.80\linewidth]{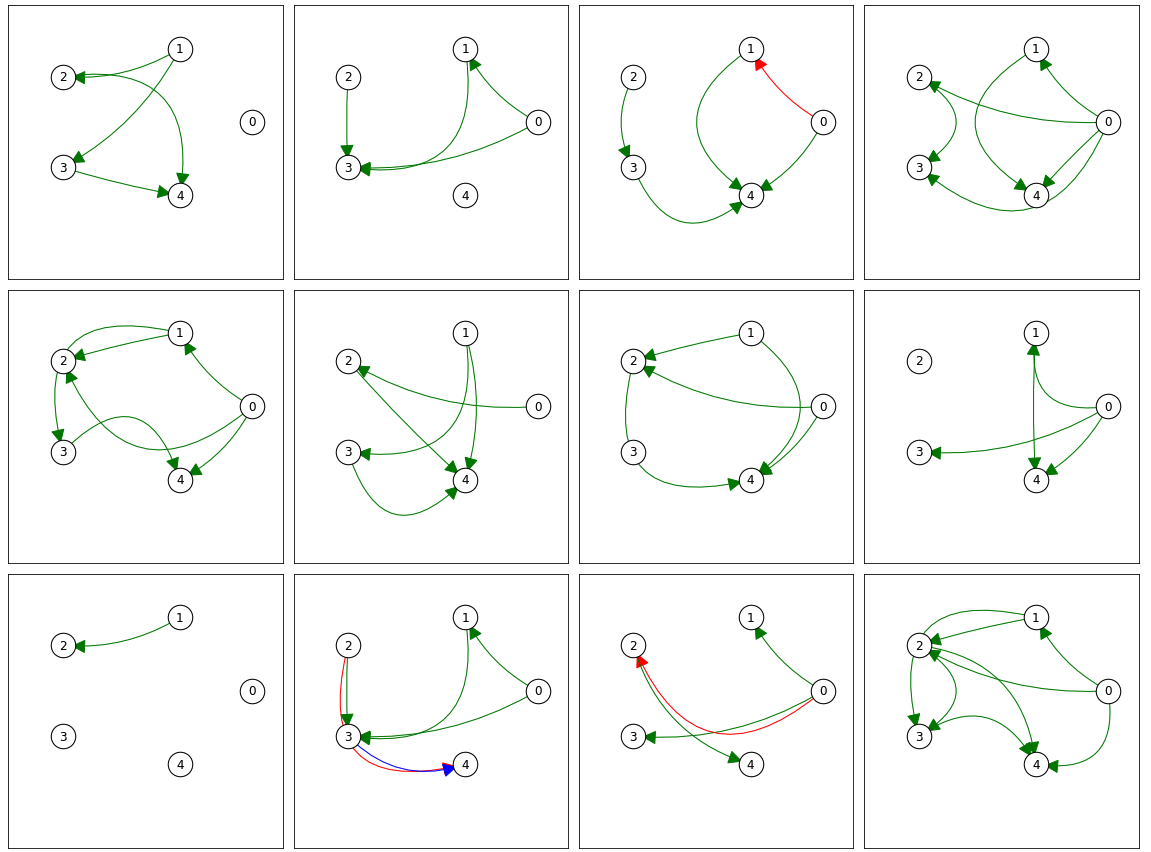}
\caption{This figures visualizes samples that our model generated on test data. The model was trained and tested on MLP data of size $5$ with \textbf{ER-$2$} graphs. The samples are randomly chosen. The green edges indicate that our model has generated the correct edges; red edges indicate edges that our model had missed; and blue edges are the ones that our model  generated, which were not in the groundtruth graph. As shown above, our model is able to generate the correct graph almost all of the times, while only occasionally generating 1 or 2 incorrect edges in a graph.}
\label{fig:samples_er2}
\end{figure}

We visualized samples that our model generated on the test data. The samples are shown in Figure \ref{fig:samples_er2} and Figure \ref{fig:samples_er1}. The samples are randomly chosen, each subplot is a sample from a distinct test data. The edges in the graph are shown in 3 colors, they each represent the following: (a) Green edges indicate that our model has generated the correct edge. (b) A red edge indicates a missing edge, that is our model did not generate the edge, which exist in the groundtruth graph. (c) A blue edge indicates a redundant edge, such that our model generated an edge that does not exist in the groundtruth graph. As shown in Figure \ref{fig:samples_er2} and \ref{fig:samples_er1}, our model is able to generate the correct graph almost all of the times.

\subsubsection{Acyclicity of generated graphs}
As discussed in Section \ref{sec:analysis}, we analyzed the generated graphs for acyclicity on Dirichlet data with $\alpha=0.1$ and $\alpha=0.25$ for graphs of size up to $N\leq20$. We evaluated the model on $128$ test datasets  per setting, and found that none of the generated graphs contains cycles, all graphs are acyclic. This is a clear indication showing that although we do not have an acyclicity regularizer, our model was still able to learn to generated DAGs.

\subsubsection{Identifiability upon seeing intervention data.}\label{sec:appendix_percent_interventions}

Our previous experiments were all performed using a fixed amount of interventions ($80\%$)  in the training and test sets. To investigate how changing the proportion of interventional data impacts the performance of our model, we train the model with varying amounts of interventions in the training set and evaluate it using different amount of interventions during test time.

To be specific, during training, the model is trained on data with varying amount of interventions,  which is randomly sampled from the uniform distribution $U[0, 0.1, 0.3, 0.5, 0.7, 0.9, 1.0]$. During test time, we evaluate the model on different amount of interventions, and we report the performance in hamming distance. We trained the model on Dirichlet data with $15$ nodes and $\alpha=0.25$. The model is trained and tested on $1000$ samples per dateset. The results are found in Figure \ref{fig:percent_intervention}. 

As shown in Figure \ref{fig:percent_intervention}, our model's performance is worst if it only receives observational data ($0\%$ interventions), and the performance of our model improves as the amount of interventional data increases. This is a clear indication that our model is able to extract information from interventional data for predicting the graph structure.

\subsubsection{Varying number of samples}\label{sec:appendix_num_samples}

\begin{wrapfigure}{r}{0.5\textwidth}
    \begin{center}
    % \vspace*{-4\baselineskip}
    \includegraphics[scale=0.42]{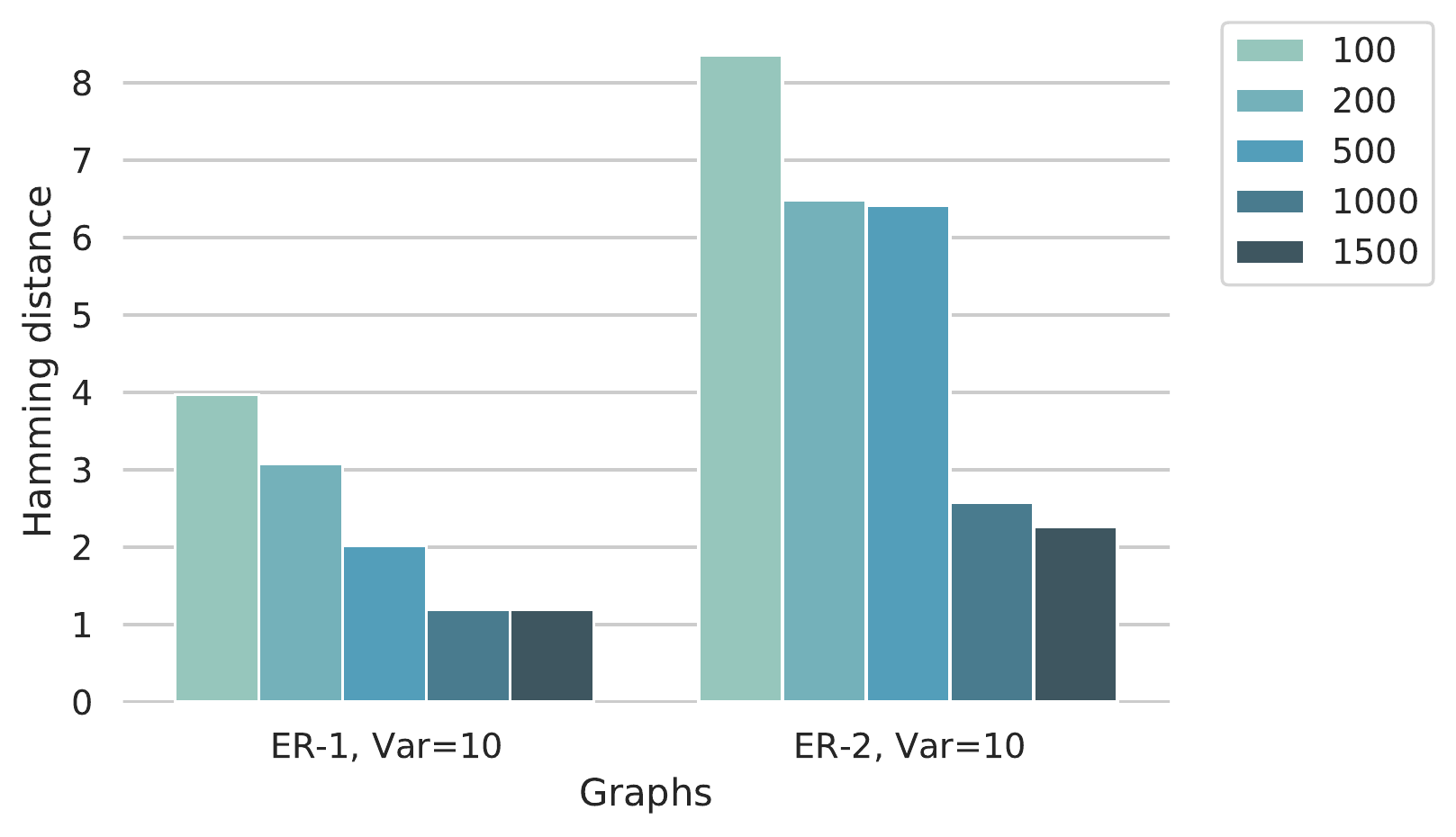}
    \caption{\textbf{Results on varying number of samples}. Hamming distance \calH~between predicted and ground-truth adjacency matrices for synthetic data. Results for \CSIVA~ trained on Dirichlet data with $N=10$ and $\alpha=0.5$ with different numbers of samples per CBNs. The model performance increases as the sample size increases. 
    }
    % \vspace{-3\baselineskip}
    \label{fig:num_samples_results}
    \end{center}
\end{wrapfigure}

We evaluated \CSIVA~on different amount of samples ($100,200,500,1000, 1500$) per CBNs. To limit the number of experiments to run, we focus on  Dirichlet data sampled from $N=10$ graphs. During  the $ith$ iteration in training,  the model takes in as input a dataset with $l_i$ samples, where $l_i$ is sampled from $U[100, 200, 500, 1000, 1500]$. During test time, we run separate evaluations on the model for test datasets with different number of samples and the results are reported in Figure~\ref{fig:num_samples_results}. We can see that the model performance improves as it observes up to $1000$ samples for ER-$1$ graphs, whereas having $1500$ samples gives slightly better results compared to $1000$ samples for ER-$2$ graphs.

\subsubsection{Ablation studies}
We investigate the role that different components play in our model, for example, what are the effects of having auxillary loss and sample-level attention for our model. The details for each experiment is described below.
\paragraph{Sample-level attention}
We first evaluated the performance of our with and without sample-level attention. In the case without sample-level attention, our model will only use node-level attention for all $10$ layers in the encoder transformer. The results are reported in Table \ref{table:sample_attention}. The models with sample-attention is a vanilla \CSIVA~model; and the model without sample-attention is \CSIVA~without sample-level attention. As one could see that the performance of the model drops significantly if sample-attention is not used. Hereby making sample-level attention a crucial component in our model.
\begin{table}[h]
{\small
\begin{tabular*}{\textwidth}{@{\extracolsep{\fill}}lcccccccc}
\toprule
 & \multicolumn{4}{c}{\textbf{ER = 1}} & \multicolumn{4}{c}{\textbf{ER = 2}} \\* \cmidrule(r){2-5} \cmidrule(l){6-9}
 & \multicolumn{2}{c}{\textbf{Var = 15}} & \multicolumn{2}{c}{\textbf{Var = 20}} & \multicolumn{2}{c}{\textbf{Var = 15}} & \multicolumn{2}{c}{\textbf{Var = 20}} \\* \cmidrule(r){2-3} \cmidrule(l){4-5} \cmidrule(l){6-7} \cmidrule(l){8-9} 
 & \multicolumn{1}{c}{\textbf{\small $\alpha=0.1$}} & 
 \multicolumn{1}{c}{\textbf{\small $\alpha=0.25$}} &
 \multicolumn{1}{c}{\textbf{\small $\alpha=0.1$}} & 
 \multicolumn{1}{c}{\textbf{\small $\alpha=0.25$}} & 
 \multicolumn{1}{c}{\textbf{\small $\alpha=0.1$}} & 
 \multicolumn{1}{c}{\textbf{\small $\alpha=0.25$}} & 
 \multicolumn{1}{c}{\textbf{\small $\alpha=0.1$}} & 
 \multicolumn{1}{c}{\textbf{\small $\alpha=0.25$}} \\
\midrule
{\small Sample} \\{\small attention} & 1.3 & 1.7 & 2.5 & 3.4 & 2.3 & 3.0 & 4.1 & 6.3 \\
\midrule
{\small No sample} \\{\small attention} & 15.3 & 15.5 & 21.8 & 22.4 & 29.3 & 29.7 & 39.4 & 40.0 \\
\bottomrule
\end{tabular*}
}
\vspace{1\baselineskip}
\caption{Results for \CSIVA~with and without \textbf{sample-level attention}. Hamming distance \calH~from learned and ground-truth edges on synthetic graphs, for known vs unknown interventions, averaged over 128 sampled graphs ($\pm$ standard deviation). The number of variables varies from 15 to 20, expected degree = 1 or 2, $\alpha$ = $0.1$ or $0.25$, and the dimensionality of the variables are fixed to 3.}
\label{table:sample_attention}
\end{table}
\paragraph{Auxiliary loss}
We also conducted ablation studies to understand the effect of the auxiliary loss in the objective function, results are reported in Table \ref{table:axu_loss}. The model with auxiliary loss is the vanilla \CSIVA~model, where as the one without is \CSIVA~trained without the auxiliary loss objective. Experiments are conducted on Dirichlet data with $\alpha=0.1$, the model does gain a small amount of performance  (\calH$<1.0$) by having the auxiliary loss, the difference becomes more apparent as the size of the graph grows, indicating that auxiliary loss plays an important role as a part of the objective function.
\begin{table*}[h]
\centering  
\begin{tabular*}{\textwidth}{@{\extracolsep{\fill}}lcccc}
\toprule
 & \multicolumn{2}{c}{\textbf{ER = 1}} & \multicolumn{2}{c}{\textbf{ER = 2}} \\* \cmidrule(r){2-3} \cmidrule(l){4-5}
 & \multicolumn{1}{c}{\textbf{Var = 15}} & \multicolumn{1}{c}{\textbf{Var = 20}} & \multicolumn{1}{c}{\textbf{Var = 15}} & \multicolumn{1}{c}{\textbf{Var = 20}} \\
\midrule
{\small Aux loss} & 1.2  & 2.4 & 2.3 & 3.4\\
{\small No aux loss} & 1.8  & 3.1 &  2.8 & 4.5\\
\bottomrule
\end{tabular*}
\caption{Results for \CSIVA~trained with and without \textbf{auxiliary loss}. Hamming distance \calH~from learned and ground-truth edges on synthetic graphs, for known vs unknown interventions, averaged over 128 sampled graphs ($\pm$ standard deviation). The number of variables varies from 15 to 20, expected degree = 1 or 2, $\alpha$ = $0.1$, and the dimensionality of the variables are fixed to 3.}
\label{table:axu_loss}
\end{table*}

\end{document}